\documentclass[11pt,]{book}
\usepackage{lmodern}
\usepackage{amssymb,amsmath}
\usepackage{ifxetex,ifluatex}
\usepackage{fixltx2e} 
\ifnum 0\ifxetex 1\fi\ifluatex 1\fi=0 
  \usepackage[T1]{fontenc}
  \usepackage[utf8]{inputenc}
\else 
  \ifxetex
    \usepackage{mathspec}
  \else
    \usepackage{fontspec}
  \fi
  \defaultfontfeatures{Ligatures=TeX,Scale=MatchLowercase}
\fi
\IfFileExists{upquote.sty}{\usepackage{upquote}}{}
\IfFileExists{microtype.sty}{%
\usepackage{microtype}
\UseMicrotypeSet[protrusion]{basicmath} 
}{}
\usepackage[margin=1in]{geometry}
\usepackage{hyperref}
\hypersetup{unicode=true,
            pdftitle={SPSA-FSR: Simultaneous Perturbation Stochastic Approximation for Feature Selection and Ranking},
            pdfauthor={Zeren D. Yenice ~zeren@advelvet.com; Niranjan Adhikari ~ndnadhikari@gmail.com; Yong Kai Wong ~yongkai1017@gmail.com; Vural Aksakalli ~vural.aksakalli@rmit.edu.au; Alev Taskin Gumus ~ataskin@yildiz.edu.tr; Babak Abbasi ~babak.abbasi@rmit.edu.au},
            pdfborder={0 0 0},
            breaklinks=true}
\usepackage{graphicx,grffile}
\makeatletter
\def\maxwidth{\ifdim\Gin@nat@width>\linewidth\linewidth\else\Gin@nat@width\fi}
\def\maxheight{\ifdim\Gin@nat@height>\textheight\textheight\else\Gin@nat@height\fi}
\makeatother
\setkeys{Gin}{width=\maxwidth,height=\maxheight,keepaspectratio}
\IfFileExists{parskip.sty}{%
\usepackage{parskip}
}{
\setlength{\parindent}{0pt}
\setlength{\parskip}{6pt plus 2pt minus 1pt}
}
\providecommand{\tightlist}{%
  \setlength{\itemsep}{0pt}\setlength{\parskip}{0pt}}
\ifx\paragraph\undefined\else
\let\oldparagraph\paragraph
\renewcommand{\paragraph}[1]{\oldparagraph{#1}\mbox{}}
\fi
\ifx\subparagraph\undefined\else
\let\oldsubparagraph\subparagraph
\renewcommand{\subparagraph}[1]{\oldsubparagraph{#1}\mbox{}}
\fi

\let\rmarkdownfootnote\footnote%
\def\footnote{\protect\rmarkdownfootnote}

\usepackage{titling}

\newcommand{\subtitle}[1]{
  \posttitle{
    \begin{center}\large#1\end{center}
    }
}

  \title{SPSA-FSR: Simultaneous Perturbation Stochastic Approximation for Feature
Selection and Ranking}
  \pretitle{\vspace{\droptitle}\centering\huge}
  \posttitle{\par}
\subtitle{An Introduction and Accuracy Performance Comparison with Other Wrapper
Methods}
  \author{Zeren D. Yenice
~\href{mailto:zeren@advelvet.com}{\nolinkurl{zeren@advelvet.com}} \\ Niranjan Adhikari
~\href{mailto:ndnadhikari@gmail.com}{\nolinkurl{ndnadhikari@gmail.com}} \\ Yong Kai Wong
~\href{mailto:yongkai1017@gmail.com}{\nolinkurl{yongkai1017@gmail.com}} \\ Vural Aksakalli\footnote{School of Science, RMIT University, Melbourne,
  Australia}
~\href{mailto:vural.aksakalli@rmit.edu.au}{\nolinkurl{vural.aksakalli@rmit.edu.au}} \\ Alev Taskin Gumus\footnote{Department of Industrial Engineering, Yildiz
  Technical University, Istanbul, Turkey}
~\href{mailto:ataskin@yildiz.edu.tr}{\nolinkurl{ataskin@yildiz.edu.tr}} \\ Babak Abbasi\footnote{School of Business IT and Logistics, RMIT
  University, Melbourne, Australia}
~\href{mailto:babak.abbasi@rmit.edu.au}{\nolinkurl{babak.abbasi@rmit.edu.au}}}
  \preauthor{\centering\large\emph}
  \postauthor{\par}
  \predate{\centering\large\emph}
  \postdate{\par}
  \date{Date: 2018-04-14}

\usepackage{lscape}
\usepackage{pdfpages}
\usepackage{amsmath}
\usepackage{algorithm}
\usepackage[noend]{algpseudocode}
\makeatletter
\def\BState{\State\hskip-\ALG@thistlm}
\makeatother

\newcommand{\blandscape}{\begin{landscape}}
\newcommand{\elandscape}{\end{landscape}}
\usepackage{hyperref}
\hypersetup{
    colorlinks = true,
    linkcolor  = {blue}
}

\let\stdsection\section
\renewcommand\section{\newpage\stdsection}

\usepackage{soul}
\usepackage{xcolor}

\definecolor{Light}{gray}{.95}
\sethlcolor{Light}

\let\OldTexttt\texttt
\renewcommand{\texttt}[1]{\OldTexttt{\hl{#1}}}

\algdef{SE}[DOWHILE]{Do}{doWhile}{\algorithmicdo}[1]{\algorithmicwhile\ #1}%

\begin{document}
\maketitle

\newpage

\textbf{Acknowledgement}

This document was written in
\href{http://rmarkdown.rstudio.com/}{Rmarkdown} inside \texttt{RStudio}
and converted with \texttt{knitr} (Version 1.17) by Xie
(\protect\hyperlink{ref-yihui}{2015}) and \texttt{pandoc}. All
visualisations in this document were created using the \texttt{ggplot2}
package (Version 2.2.1) by Wickham
(\protect\hyperlink{ref-hadley}{2016}) in \texttt{R} 3.4.1 (R Core Team
\protect\hyperlink{ref-R}{2017}).

\newpage

\tableofcontents

\newpage

\chapter{Introduction}\label{introduction}

The recent data storage technology has enabled improved data collection
and capacity which lead to more massive volumes of data collected and
higher data dimensionality. Therefore, searching an optimal set of
predictive features within a noisy dataset became an indispensable
process in supervised machine learning to extract useful patterns.
Excluding irrelevant and redundant features for problems with large and
noisy datasets provides the following benefits: (1) reduction in
overfitting, (2) better generalisation of models, (3) superior
prediction performance, and (4) CPU and memory efficient and fast
prediction models. There are two major techniques for reducing data
dimensionality: Feature Selection (FS) and Feature Extraction (FE).

The core principle of Feature extraction (FE) is to generate new
features by usually combining original ones then maps acquired features
into a new (lower dimensional) space. The well-known FE techniques are
Principle Component Analysis (PCA), Linear Discriminant Analysis (LDA)
and ISO-Container Projection (ISOCP) (Z. Zheng, Chenmao, and Jia
\protect\hyperlink{ref-iso}{2010}).The most critical drawback of the FE
method is that the reduced set of newly generated features might lose
the interpretability of the original dataset. Relating new features to
original features proves difficult in FE, and consequently, any further
analysis of transformed features become limited.

Feature selection (FS) is loosely defined as selecting a subset of
available features in a dataset that is associated with the response
variable by excluding irrelevant and redundant features (Aksakalli and
Malekipirbazari \protect\hyperlink{ref-vural}{2016}). In contrast with
the FE method, the FS preserves the physical meaning of original
features and provides better readability and interpretability. Suppose
the dataset has \(p\) original features, the FS problem has \(2^p-1\)
possible solutions and thus it is an NP-hard problem due to an
exponential increase in computational complexity with \(p\). FS methods
can fall into four categories: wrapper, filter, embedded and hybrid
methods.

A wrapper method uses a predictive score of given learning algorithm to
appraise selected features. While wrappers usually give the best
performing set of features, they are very computationally intensive as
they train a new model for each subset. In contrast with wrapper
methods, filter methods evaluate features without utilising any learning
algorithms. Filter methods measure statistical characteristics of data
such as correlation, distance and information gain to eliminate
insignificant features. The commonly used filter methods include Relief
(Sikonia and Kononenko \protect\hyperlink{ref-sikonia}{2003}) and
information-gain based models (H. Peng, Long, and Ding
\protect\hyperlink{ref-peng}{2005}). Senawi, Wei, and Billings
(\protect\hyperlink{ref-senawi}{2017}) proposed a filter method that
applies straightforward hill-climbing search and correlation
characteristics such as conditional variance and orthogonal projection.
Bennasar, Hicks, and Setchi (\protect\hyperlink{ref-bennasar}{2015})
introduced two filter methods which aim to mitigate the issue of the
overestimated feature significance by using mutual information and the
maximin criterion.

Embedded methods aim to improve a predictive score like wrapper methods
but carry out FS process in the learning time
(\protect\hyperlink{ref-cadenas}{2013}). Their computational complexity
tends to fall between filters and wrappers. Some of the popular embedded
methods are Least Absolute Shrinkage and Selection Operator (LASSO)
(Tibshirani \protect\hyperlink{ref-lasso}{1996}), Support Vector Machine
recursive feature elimination (SVM-RFE) (Guyon et al.
\protect\hyperlink{ref-guyon}{2002}) and Random-Forest (Leo
\protect\hyperlink{ref-rf}{2001}). Hybrid methods are a combination of
filter and wrapper methods. Hsu, Hsieh, and Ming-Da
(\protect\hyperlink{ref-hsu}{2011}) combined F-score and information
gain with the sequential forward and backward searches to solve
bioinformatics problems. Cadenas, Garrido, and Martinez
(\protect\hyperlink{ref-cadenas}{2013}) presented the blending of Fuzzy
Random Forest and discretisation process (filter). Apolloni, Leguizamón,
and Alba (\protect\hyperlink{ref-apolloni}{2016}) developed two hybrid
algorithms which are a combination of rank-based filter FS method and
Binary Differential Evolution (BDE) algorithm. Furthermore,
MIMAGA-Selection method (Lu et al. \protect\hyperlink{ref-lu}{2017}) is
a mix of the adaptive genetic algorithm (AGA) and mutual information
maximisation (MIM).

A wrapper method typically begins subsetting the feature set, evaluating
the set by the performance criterion of the classifier, and repeating
until the desired quality is obtained (Kohavi and John
\protect\hyperlink{ref-kohavi}{1997}). Wrapper FS algorithms are
classified into four categories: complete, heuristic and meta-heuristic
search, and optimisation-based. As complete search algorithms become
infeasible with increasing number of features, they are not appropriate
for big-data problems (L. Wang, Wang, and Chang
\protect\hyperlink{ref-wang2}{2016}). Heuristic algorithms search good
local optima before barren subsets. Besides, a suitable heuristic search
algorithm could determine a global optimum given sufficient computation
time. Heuristic wrapper methods contain branch-and-bound techniques,
beam search, best-first and greedy hill-climbing algorithms. Prominent
greedy hill-climbing algorithms are Sequential Forward Search (which
starts with an empty feature set and most significant features are
gradually added to the feature set) and Sequential Backward Search
(start with a full feature set and gradually eliminate insignificant
features). However, these models do not re-evaluate and unwittingly
exclude the eliminated features which might be possibly predictive in
another feature sets (Guyon and Elisseeff
\protect\hyperlink{ref-guyon2}{2003}). Thus, Sequential Forward Floating
Selection (SFFS) and Sequential Backward Floating Selection (SBFS)
methods are developed to mitigate deal with the early elimination issue
(Pudil, Novovicová, and Kittler \protect\hyperlink{ref-pudil}{1994}).
For instance, whenever a feature is added to the feature set, the SFFS
method involves checking the feature set by removing any worse feature
if the conditions are satisfied. Such process can correct wrong
decisions made in the previous steps.

Meta-heuristics are problem-independent methods, and so they can surpass
complete search and heuristic sequential feature selection methods.
However, meta-heuristic algorithms might be computationally expensive.
They do not guarantee optimality for resulted feature sets and require
additional parameter tuning in order to provide more robust results.
Several meta-heuristic methods have been applied to FS problems are
genetic algorithms (GA) (Oluleye, Armstrong, and Diepeveen
\protect\hyperlink{ref-oluleye}{2014}), (Raymer et al.
\protect\hyperlink{ref-raymer}{2000}), (Tsai, Eberle, and Chu
\protect\hyperlink{ref-tsai}{2013}), ant colony optimization (Al-Ani
\protect\hyperlink{ref-ani}{2005}), (Wan et al.
\protect\hyperlink{ref-wan}{2006}), binary flower pollination (Sayed,
Nabil, and Badr \protect\hyperlink{ref-sayed}{2016}), simulated
annealing (Debuse and Rayward-Smith
\protect\hyperlink{ref-debuse}{1997}), forest optimization (Ghaemi and
Feizi-Derakhshi \protect\hyperlink{ref-ghaemi}{2016}), tabu search
(Tahir, Bouridane, and Kurugollu \protect\hyperlink{ref-tahir}{2007}),
bacterial foraging optimization (Y. Chen et al.
\protect\hyperlink{ref-chen}{2017}), particle swarm optimization (X.
Wang et al. \protect\hyperlink{ref-wang3}{2007}), binary black hole
(Pashaei and Aydin \protect\hyperlink{ref-aydin}{2017}) and hybrid whale
optimization (Mafarja and Mirjalili
\protect\hyperlink{ref-mafarja}{2017}).

Optimisation-based methods treat the FS task as a mathematical
optimisation problem. One of the optimisation-based methods is Nested
Partitions Method (NP) introduced by (Ólafsson and Yang
\protect\hyperlink{ref-olafsson}{2005}). The NP method randomly searches
the entire space of possible feature subsets by partitioning the search
space into regions. It analyses each regional result and then aggregates
them to determine the search direction. Another optimisation-based
method is the Binary Simultaneous Perturbation Stochastic Approximation
algorithm (BSPSA) (Aksakalli and Malekipirbazari
\protect\hyperlink{ref-vural}{2016}). BSPSA simultaneously approximates
the gradient of each feature and eventually determine a feature subset
that yields the best performance measurement. Although it outperforms
most of the wrapper algorithms, its computational cost is higher.

This study introduces the Simultaneous Perturbation Stochastic
Approximation for Feature Selection (SPSA-FS) algorithm which improves
BSPSA using non-monotone Barzilai \& Borwein (BB) search method. The
SPSA-FS algorithm improves BSPSA to be computationally faster and more
decisive by incorporating the following:

\begin{enumerate}
\def\labelenumi{\arabic{enumi}.}
\tightlist
\item
  Non-monotone step size calculated via the BB method,
\item
  Averaging of \(n\) number of gradient approximation, and
\item
  \(m\) period gain smoothing.
\end{enumerate}

The rest of this document is organised as follow. Chapter \ref{section2}
gives an overview of both SPSA and BSPSA algorithms and a simple
two-iteration computational example to illustrate its concept. Chapter
\ref{section3} introduces the SPSA-FS algorithm which utilises the BB
method to mitigate the slow convergence of BSPSA at a cost of minimal
decline in performance accuracy. With SPSA-FS as a benchmark, Chapter
\ref{section4} compares the performance of other wrappers in various
classification and regression problems using the open datasets. Chapter
\ref{section5} concludes.

\chapter{Background}\label{section2}

\section{Stochastic Pseudo-Gradient Descent
Algorithms}\label{stochastic-pseudo-gradient-descent-algorithms}

Introduced by J. Spall (\protect\hyperlink{ref-spall}{1992}), SPSA is a
pseudo-gradient descent stochastic optimisation algorithm. It starts
with a random solution (vector) and moves toward the optimal solution in
successive iterations in which the current solution is perturbed
simultaneously by random offsets generated from a specified probability
distribution.

Let \(\mathcal{L}: \mathbb{R}^{p} \mapsto \mathbb{R}\) be a real-valued
objective function. The gradient descent approach startes searching for
a local minimum of \(\mathcal{L}\) with an initial guess of a solution.
Then, it evaluates the gradient of the objective function i.e.~the
first-order partial derivatives \(\nabla \mathcal{L}\), and moves in the
direction of \(-\nabla \mathcal{L}\). The gradient descent algorithm
attempts to converge to a local optima where the gradient is zero. In
the context of a supervised machine learning problem, \(\mathcal{L}\) is
sometimes known as a ``loss'' function as a case of minimisation
problem.

However, the gradient descent approach is not applicable in the
situations where the loss function is not explicitly known and so its
gradient. Here come stochastic pseudo-gradient descent algorithms to
rescue. These algorithms, including \textbf{SPSA}, approximate the
gradient from noisy loss function measurements and hence do not need the
information about the (unobserved) functional form of the loss function.

\section{SPSA Algorithm}\label{spsa-algorithm}

Given \(w \in D \subset \mathbb{R}^{p}\), let
\(\mathcal{L}(w): D \mapsto \mathbb{R}\) be the loss function where its
functional form is unknown but one can observe noisy measurement:

\begin{center}
  \begin{align}
  y(w) & := \mathcal{L}(w) + \varepsilon(w) \label{lossfunction}
  \end{align}
\end{center}

where \(\varepsilon\) is the noise and \(y\) is the noise measurement.
Let \(g(w)\) denote the gradient of \(\mathcal{L}\):

\[g(w): = \nabla \mathcal{L} = \frac{\partial \mathcal{L} }{\partial w}\]
SPSA starts with an initial solution \(\hat{w}_0\) and iterates
following the recursion below in search for a local minima \(w^{*}\):

\[\hat{w}_{k+1} := \hat{w}_{k} - a_k \hat{g}(\hat{w}_{k})\]

where:

\begin{itemize}
\tightlist
\item
  \(a_{k}\) is an iteration gain sequence; \(a_{k} \geq 0\), and
\item
  \(\hat{g}(\hat{w}_{k})\) is the approximate gradient at
  \(\hat{w}_{k}\).
\end{itemize}

Let \(\Delta_k \in \mathbb{R}^p\) be a \textbf{simultaneous perturbation
vector} at iteration \(k\). SPSA imposes certain regularity conditions
on \(\Delta_k\) (J. Spall \protect\hyperlink{ref-spall}{1992}):

\begin{itemize}
\tightlist
\item
  The components of \(\Delta_{k}\) must be mutually independent,
\item
  Each component of \(\Delta_{k}\) must be generated from a symmetric
  zero mean probability distribution,
\item
  The distribution must have a finite inverse, and
\item
  \(\{ \Delta_k\}_{k=1}\) must be a mutually independent sequence which
  is independent of \(\hat{w}_0, \hat{w}_1,...\hat{w}_k\).
\end{itemize}

The finite inverse requirement precludes \(\Delta_k\) from uniform or
normal distributions. A good candidate is a symmetric zero mean
Bernoulli distribution, say \(\pm 1\) with 0.5 probability. SPSA
``perturbs'' the current iterate \(\hat{w}_k\) by an amount of
\(c_k \Delta_k\) in each direction of \(\hat{w}_k + c_k \Delta_k\) and
\(\hat{w}_k - c_k \Delta_k\) respectively. Hence, the
\textbf{simultaneous perturbations} around \(\hat{w}_{k}\) are defined
as:

\[\hat{w}^{\pm}_k := \hat{w}_{k} \pm  c_k \Delta_k\] where \(c_k\) is a
nonnegative gradient gain sequence. The noisy measurements of
\(\hat{w}^{\pm}_k\) at iteration \(k\) become:

\[y^{+}_k:=\mathcal{L}(\hat{w}_k + c_k \Delta_k) + \varepsilon_{k}^{+}\]
\[y^{-}_k:=\mathcal{L}(\hat{w}_k - c_k \Delta_k) + \varepsilon_{k}^{-}\]

where
\(\mathbb{E}( \varepsilon_{k}^{+} - \varepsilon_{k}^{-}|\hat{w}_0, \hat{w}_1,...\hat{w}_k, \Delta_k) = 0 \forall k\).
Therefore, \(\hat{g}_k\) is computed as:

\[\hat{g}_k(\hat{w}_k):=\bigg[ \frac{y^{+}_k-y^{-}_k}{w^{+}_{k1}-w^{-}_{k1}},...,\frac{y^{+}_k-y^{-}_k}{w^{+}_{kp}-w^{-}_{kp}} \bigg]^{T} = \bigg[ \frac{y^{+}_k-y^{-}_k}{2c_k \Delta_{k1}},...,\frac{y^{+}_k-y^{-}_k}{2c_k \Delta_{kp}} \bigg]^{T} = \frac{y^{+}_k-y^{-}_k}{2c_k}[\Delta_{k1}^{-1},...,\Delta_{kp}^{-1}]^{T}\]
At each iteration \(k\), SPSA evaluates \textbf{three noisy measurements
of loss function}: \(y^{+}_k\), \(y^{-}_k\), and \(y(\hat{w}_{k+1})\).
\(y^{+}_k\) and \(y^{-}_k\) are used to approximate the gradient whereas
\(y(\hat{w}_{k+1})\) is used to measure the performance of next iterate,
\(\hat{w}_{k+1}\). J. C. Spall (\protect\hyperlink{ref-spall2}{2003})
states that if certain conditions hold, \(w_k \mapsto w^{*}\) as
\(k \rightarrow \infty\). See J. C. Spall
(\protect\hyperlink{ref-spall2}{2003}) for more information about
theoretical aspects of SPSA.

J. Spall (\protect\hyperlink{ref-spall}{1992}) proposed the following
functions for tuning parameter

\begin{center}
  \begin{align}
  a_k & := \frac{a}{(A+k)^{\alpha}} \label{spsaStepSize} \\
  c_k & := \frac{c}{\gamma^{k}}
  \end{align}
\end{center}

\(A\), \(a\), \(\alpha\), \(c\) and \(\gamma\) are pre-defined; these
parameters must be fine-tuned properly. SPSA does not have automatic
stopping rules. So, we can specify a maximum number of iterations as a
stopping criterion. In addition, the iteration sequence
\ref{spsaStepSize} must be monotone and satisfy:

\[\lim_{k\rightarrow\infty} a_k = 0\]

\section{Binary SPSA Algorithm}\label{bspsa}

J. Spall and Qi (\protect\hyperlink{ref-wang}{2011}) provided a discrete
version of SPSA where \(w \in \mathbb{Z}^{p}\). Therefore, a binary
version of SPSA (BSPSA) is a special case of the discrete SPSA with
fixed perturbation parameters. The loss function becomes
\(\mathcal{L}: \{0,1\}^{p} \mapsto \mathbb{R}\). BSPSA is different from
the conventional SPSA in two ways:

\begin{enumerate}
\def\labelenumi{\arabic{enumi}.}
\tightlist
\item
  The gain sequence \(c_k\) is constant, \(c_k=c\);
\item
  \(\hat{w}_k^{\pm}\) are bounded and rounded before \(y_k^{\pm}\) are
  evaluated.
\end{enumerate}

Algorithm \ref{pseudoCodes} illustrates the pseudo code for BSPSA
Algorithm.

\begin{algorithm}
\caption{BSPSA Algorithm} \label{pseudoCodes}
\begin{algorithmic}[1]
\Procedure{\underline{BSPSA}($\hat{w}_0$, $a$, $A$, $\alpha$,  $c$, $M$)}{}
\BState Initialise $k = 0$
\BState \textbf{do}:
\State Simulate $\Delta_{k, j} \sim \text{Bernoulli}(-1, +1)$ with $\mathbb{P}(\Delta_{k, j}=1) = \mathbb{P}(\Delta_{k, j}=-1) = 0.5$ for $j=1,..p$
\State $\hat{w}^{\pm}_k = \hat{w}_{k} \pm c \Delta_k$
\State $\hat{w}^{\pm}_k = B(\hat{w}^{\pm}_k)$ \Comment{$B( \bullet)$ = component-wise $[0,1]$ operator }
\State $\hat{w}^{\pm}_k = R(\hat{w}^{\pm}_k)$ \Comment{$R( \bullet)$ = component-wise rounding operator}
\State $y^{\pm}_k =\mathcal{L}(\hat{w}_k \pm c_k \Delta_k) \pm \varepsilon_{k}^{\pm}$
\State $\hat{g}_k(\hat{w}_k) =\bigg( \frac{y^{+}_k-y^{-}_k}{2c}\bigg)[\Delta_{k1}^{-1},...,\Delta_{kp}^{-1}]^{T}$ \Comment{$\hat{g}_k(\hat{w}_k)$ = the gradient estimate}
\State $\hat{w}^{\pm}_k = \hat{w}_{k} \pm  a_k \hat{g}_k(\hat{w}_k)$ \Comment{$a_k = \frac{A}{(a+k)^{\alpha}}$}
\State $k = k + 1$
\BState \textbf{while} ($k < M$)
\BState \textbf{Output}: $\hat{w}^{\pm}_M = R(\hat{w}^{\pm}_M)$
\EndProcedure
\end{algorithmic}
\end{algorithm}

\section{Illustration of BSPSA Algorithm in Feature
Selection}\label{fbspsa}

Let \textbf{X} be \(n \times p\) data matrix of \(p\) features and \(n\)
observations whereas \(Y\) denotes the \(n \times 1\) response vector.
Do not confuse with \(y\) which represents the functional form of the
loss function (see Equation \ref{lossfunction}).
\(\{\)\textbf{X}\(, Y \}\) consitute as the dataset. Let
\(X:= \{ X_1, X_2, ....X_p \}\) denote the feature set where \(X_j\)
represents the \(j^{th}\) feature in \textbf{\(X\)}. For a nonempty
subset \(X' \subset X\), we define \(\mathcal{L}_{C}(X', Y)\) as the
true value of performance criterion of a wrapper classifier (the model)
\(C\) on the dataset. As \(\mathcal{L}_{C}\) is not known, we train the
classifier \(C\) and compute the error rate, which is denoted by
\(y_C(X', Y)\). Therefore, \(y_C = \mathcal{L}_C + \varepsilon\). The
wrapper FS problem is defined as determining the non-empty feature set
\(X^{*}\):

\[X^{*} := \arg \min_{X' \subset X}y_C(X', Y)\]

It would be the best to use some examples to illustrate how Binary SPSA
method works. With a block diagram (\protect\hyperlink{ref-vural}{2016},
Figure 1, p.~6), Aksakalli and Malekipirbazari
(\protect\hyperlink{ref-vural}{2016}) provided an example with one
iteration and a hypothetical dataset with four features. In this
section, for completeness, we show the next example depicts how the SPSA
algorithm runs in two iterations. Suppose we have:

\begin{itemize}
\tightlist
\item
  6 features i.e. \(p=\) 6;
\item
  \(y_C\) as a cross-validated error rate of a classifer;
\item
  Parameters: \(c=\) 0.05, \(a=\) 0.75, \(A=\) 100, and \(\alpha=\) 0.6;
  and
\item
  Maximum 2 iterations i.e. \(M=\) 2
\item
  An initial guess \(\hat{w}_0:=\) {[}0.5, 0.5, 0.5, 0.5, 0.5, 0.5{]}.
\end{itemize}

\newpage

At the \textbf{first} iteration, i.e. \(k=\) 0;

\begin{enumerate}
\def\labelenumi{\arabic{enumi}.}
\tightlist
\item
  Generate \(\Delta_0\) as {[}-1, -1, 1, 1, -1, 1{]} from a Bernoulli
  Distribution.
\item
  Compute \(\hat{w}^{\pm}_0 = \hat{w}_{0} \pm c \Delta_0\)
\end{enumerate}

\begin{center}
  \begin{align*}
    \hat{w}^{+}_0 & =\hat{w}_{0} + c \Delta_0 = [0.45, 0.45, 0.55, 0.55, 0.45, 0.55]\\
    \hat{w}^{-}_0 & =\hat{w}_{0} - c \Delta_0 = [0.55, 0.55, 0.45, 0.45, 0.55, 0.45]
  \end{align*}
\end{center}

\begin{enumerate}
\def\labelenumi{\arabic{enumi}.}
\setcounter{enumi}{2}
\tightlist
\item
  Bound \(w^{\pm}_0\):
\end{enumerate}

\begin{center}
  \begin{align*}
    w^{+}_0 & = B(w^{+}_0)= [0.45, 0.45, 0.55, 0.55, 0.45, 0.55]\\
    w^{-}_0 & = B(w^{-}_0)= [0.55, 0.55, 0.45, 0.45, 0.55, 0.45]   
  \end{align*}
\end{center}

\begin{enumerate}
\def\labelenumi{\arabic{enumi}.}
\setcounter{enumi}{3}
\tightlist
\item
  Round \(w^{\pm}_0\)
\end{enumerate}

\begin{center}
  \begin{align*}
    w^{+}_0 & = R(w^{+}_0)= [0, 0, 1, 1, 0, 1]\\
    w^{-}_0 & = R(w^{-}_0)= [1, 1, 0, 0, 1, 0]   
  \end{align*}
\end{center}

\begin{enumerate}
\def\labelenumi{\arabic{enumi}.}
\setcounter{enumi}{4}
\item
  Evaluate \(y^{+}_{1} := y(\){[}0, 0, 1, 1, 0, 1{]}\()\) and
  \(y^{-}_{1} := y(\){[}1, 1, 0, 0, 1, 0{]}\()\). Assume \(y^{+}_{1} =\)
  0.32 and \(y^{-}_{1} =\) 0.53.
\item
  Compute
  \(\hat{g}_0 (\hat{w}_0):=\bigg(\frac{y_1^{+}-y_1^{-}}{2c} \bigg) \Delta_0^{-1} =\)
  {[}2.1, 2.1, -2.1, -2.1, 2.1, -2.1{]}.
\item
  Calculate \(a_0=\frac{a}{(100+0)^\alpha}=\) 0.047. Compute
  \(\hat{w}_1=\hat{w}_0-a_0\hat{g}_0 (\hat{w}_0)\)

  \begin{center}
    \begin{align*}
  \hat{w}_1 & = [0.5, 0.5, 0.5, 0.5, 0.5, 0.5] - 0.047[2.1, 2.1, -2.1, -2.1, 2.1, -2.1]\\
            & = [0.4013, 0.4013, 0.5987, 0.5987, 0.4013, 0.5987]. 
    \end{align*}
  \end{center}
\end{enumerate}

\newpage

At the \textbf{second} iteration, i.e. \(k=\) 1;

\begin{enumerate}
\def\labelenumi{\arabic{enumi}.}
\tightlist
\item
  Generate \(\Delta_1\) as {[}-1, 1, 1, -1, 1, 1{]} from a Bernoulli
  Distribution.
\item
  Compute \(\hat{w}^{\pm}_1 = \hat{w}_{1} \pm c \Delta_1\)
\end{enumerate}

\begin{center}
  \begin{align*}
    \hat{w}^{+}_1 & =\hat{w}_{1} + c \Delta_1 = [0.351, 0.451, 0.649, 0.549, 0.451, 0.649]\\
    \hat{w}^{-}_1 & =\hat{w}_{1} - c \Delta_1 = [0.451, 0.351, 0.549, 0.649, 0.351, 0.549]
  \end{align*}
\end{center}

\begin{enumerate}
\def\labelenumi{\arabic{enumi}.}
\setcounter{enumi}{2}
\tightlist
\item
  Bound \(w^{\pm}_1\):
\end{enumerate}

\begin{center}
  \begin{align*}
    w^{+}_1 & = B(w^{+}_1)= [0.351, 0.451, 0.649, 0.549, 0.451, 0.649]\\
    w^{-}_1 & = B(w^{-}_1)= [0.451, 0.351, 0.549, 0.649, 0.351, 0.549]  
  \end{align*}
\end{center}

\begin{enumerate}
\def\labelenumi{\arabic{enumi}.}
\setcounter{enumi}{3}
\tightlist
\item
  Round \(w^{\pm}_0\)
\end{enumerate}

\begin{center}
  \begin{align*}
    w^{+}_1 & = R(w^{+}_1)= [0, 0, 1, 1, 0, 1]\\
    w^{-}_1 & = R(w^{-}_1)= [0, 0, 1, 1, 0, 1]   
  \end{align*}
\end{center}

\begin{enumerate}
\def\labelenumi{\arabic{enumi}.}
\setcounter{enumi}{4}
\item
  Evaluate \(y^{+}_{2} := y(\){[}0, 0, 1, 1, 0, 1{]}\()\) and
  \(y^{-}_{2} := y(\){[}0, 0, 1, 1, 0, 1{]}\()\). Assume \(y^{+}_{2} =\)
  0.53 and \(y^{-}_{2} =\) 0.38.
\item
  Compute
  \(\hat{g}_1 (\hat{w}_1):=\bigg(\frac{y_2^{+}-y_2^{-}}{2c} \bigg) \Delta_1^{-1} =\)
  {[}-1.5, 1.5, 1.5, -1.5, 1.5, 1.5{]}.
\item
  Calculate \(a_1=\frac{a}{(100+1)^\alpha}=\) 0.047. Compute
  \(\hat{w}_2=\hat{w}_1-a_1\hat{g}_1 (\hat{w}_1)\)

  \begin{center}
    \begin{align*}
  \hat{w}_2 & = [0.401, 0.401, 0.599, 0.599, 0.401, 0.599] - 0.047[-1.5, 1.5, 1.5, -1.5, 1.5, 1.5]\\
            & = [0.471, 0.33, 0.529, 0.67, 0.33, 0.529]. 
    \end{align*}
  \end{center}
\end{enumerate}

In the \textbf{final} step, let's round \(\hat{w}_2\) to the solution
vector of {[}0, 0, 1, 1, 0, 1{]}. This means the best performing feature
set include features 3, 4, 6.

\chapter{SPSA-FS Algorithm}\label{section3}

\section{Barzilai-Borwein (BB) Method}\label{barzilai-borwein-bb-method}

The philosophy of the non-monotone methods involves remembering the data
provided by previous iterations. As the first non-monotone search
method, the Barzilai-Borwein method (Barzilai and Borwein
\protect\hyperlink{ref-bb}{1988}) is described as the gradient method
with two point-step size. Motivated by Newton's method, the BB method
aims to approximate the Hessian matrix, instead of direct computation.
In other words, it computes a sequence of objective values that are not
monotonically decreasing. Proven by Barzilai and Borwein
(\protect\hyperlink{ref-bb}{1988}), the BB method significantly
outperforms the classical steppest decent method by means of better
performances and lower computation costs.

Consider an unconstrained optimisation problem expressed in Equation
\ref{3.1}:

\begin{center}
  \begin{align}
    \min_{x\epsilon{R^p}} f(x) \label{3.1}
  \end{align}
\end{center}

The steepest decent method, also known as the Cauchy method (Cauchy
\protect\hyperlink{ref-cauchy}{1847}), uses the negative of the gradient
as the search direction to locate next point with step size determined
by exact or backtracking line search. The next point in the search
direction is given by

\[x_{k+1} = x_{k} + \alpha_{k}d_{k}\]

The negative gradient of \(f\) at \(x_{k}\) is defined as:

\[d_{k} = -\nabla f(x_{k})\]

The step size \(\alpha_{k}\) is defined as:

\begin{center}
  \begin{align}
    \alpha_k = \arg \min_\alpha f(x_k + \alpha d_k) \label{3.2}
  \end{align}
\end{center}

The gradient will denoted as \(g_k = g(x_k) = \nabla f(x_k)\). Suppose
\(f\) has a quadratic form such that:

\[f(x) = \frac{1}{2} x^T Qx - b^Tx + c\]

then the exact line search (\ref{3.2}) becomes explicit and the stepsize
\(\alpha_k\) of steepest decent method can be derived as:

\begin{center}
  \begin{align}
    \alpha_k = \frac{g_k^Tg_k}{g_k^T Qg_k} \label{3.3}
  \end{align}
\end{center}

Although Cauchy's method is simple and uses optimal property
(\ref{3.2}), it does not use the second order information. As a result,
it tends to perform poorly and suffers from the ill-conditioning
problem. The alternative is Newton's Method (Nocedal and Wright
\protect\hyperlink{ref-no}{2006}, 44--46), which finds the next trial
point by

\[x_{k+1} = x_k - (F_{k})^{-1} g_k\]

where \(F_k = \nabla^2 f(x_k)\) which is computationally very expensive
and sometimes requires modification if \(F_k \nsucc 0\). On the other
hand, the BB method choose the step size \(\alpha_k\) by solving either
of following least squares problems so that \(\alpha_k g_k\)
approximates \((F_k)^{-1}g_k\)

\begin{center}
  \begin{align}
    & \min_\alpha ||\nabla{x} - \alpha\nabla{g}||^2 \label{eqn3.4} \\
    & \min_\alpha ||\alpha\nabla{x} - \nabla{g}||^2 \label{eqn3.5}
  \end{align}
\end{center}

where \(\nabla x = x_k - x_{k-1}\), and \(\nabla g = g_k - g_{k-1}\).
The respective solutions to Problem \ref{eqn3.4} and Problem
\ref{eqn3.5} are:

\begin{center}
  \begin{align}
    & \alpha_k = \frac{\nabla{x^T}\nabla{g}}{\nabla{g^T}\nabla{g}} \label{eqn3.6} \\
    & \alpha_k = \frac{\nabla x^T\nabla x} {\nabla x^T\nabla g} \label{eqn3.7}
  \end{align}
\end{center}

Raydan (\protect\hyperlink{ref-raydan}{1993}), Molina and Raydan
(\protect\hyperlink{ref-molina}{1996}), and Y. Dai and Liao
(\protect\hyperlink{ref-dai2}{2002}) studied the convergence analysis of
the BB method and found that the BB method BB linearly converges in a
strictly convex quadratic form. In the literature, the famous BB methods
include Caughy BB and Cyclic BB. Cauchy BB (Raydan and Svaiter
\protect\hyperlink{ref-raydan2}{2002}) combines BB and Cauchy method and
reduces computational work by half compared to BB. Cauchy method
outperforms BB for quadratic problems when \(g_k\) is not almost an
eigenvector of \(Q\). Meanwhile, Cyclic BB Method (Y. Dai et al.
\protect\hyperlink{ref-dai}{2006}) involves specifying a predetermined
cycle length and uses the same calculated step size \ref{eqn3.6} until
the cycle length is reached before proceeding to the next step size.
Therefore, the computation time is very sensitive to the choice of cycle
length affects. Furthermore, Cauchy BB includes steppest descent method
whereas Cyclic BB has an extra process which determines the appropriate
cyclic length. Given these shortcomings, the original BB method with a
smoothing effect is implemented in the SPSA-FS algorithm.

\section{BB Method in SPSA-FS}\label{bb-method-in-spsa-fs}

Since it relies on a monotone step size \(a_k\), the BSPSA algorithm has
a slow convergence rate, which renders its usefulness in a time-critical
situation. The slow convergence issue become more acute in the larger
data size. To reduce the convergence time, we propose the nonmonotone BB
method.

It is important to notice the difference in the notation to represent
the step size in the literature. For the BB method, it is typically
denoted by \(\alpha_k\) while it is \(a_k\), which is also known as the
iteration gain sequence, in SPSA (see Equation \ref{spsaStepSize}). To
be consistent with BSPSA, we shall modify the latter \(a_k\) to
\(\hat{a_{k}}\) and express the BB method's step size \ref{eqn3.6} as:

\begin{center}
  \begin{align}
    \hat{a}_k &= \frac{\nabla{\hat{w}}^T\nabla{\hat{g}(\hat{w})}}{\nabla{\hat{g}^T(\hat{w})}\nabla{\hat{g}(\hat{w})}} \label{eqn3.8}
  \end{align}
\end{center}

We use \(\hat{a_{k}}\) to indicate it is an estimate rather than a
closed form like Equation \ref{spsaStepSize}. Sometimes the gain can be
negative such that \(\nabla{\hat{w}^T}\nabla{\hat{g}(\hat{w})} < 0\).
This is possible because the Hessian of \(f\) might include negative
eigenvalues at \(\nabla{\hat{w}}\) i.e.~a point between \(\hat{w_k}\)
and \(\hat{w_{k-1}}\) (Y. Dai et al. \protect\hyperlink{ref-dai}{2006}).
Consequently, it is necessary to set closed boundaries around the gain
to ensure it is monotonic. Therefore, the current gain (equation
\ref{eqn3.8}) becomes:

\begin{center}
  \begin{align}
    \hat{a}_k^{'} &= \max\{a_{\min},\min\{\hat{a}_k,a_{\max}\}\} \label{eqn3.9}
  \end{align}
\end{center}

where \(a_{\min}\) and \(a_{\max}\) are the minimum and the maximum of
gain sequence \(\{\hat{a}_k\}_{k}\) at the current iteration \(k\)
respectively.

\textbf{Gain Smoothing}

Tan et al. (\protect\hyperlink{ref-tan}{2016}) propose to smooth the
gain as the following:

\begin{center}
  \begin{align}
  \hat{b}_k = \frac{\sum_{n=k-t}^k{\hat{a}_{n}^{'}}}{t+1} \label{eqn3.10}
  \end{align}
\end{center}

The role of \(\hat{b}_k\) is to eliminate the irrational fluctuations in
the gains and ensure the stability of the SPSA-FS algorithm. SPSA-FS
averages the gains at the current and last two iterations, i.e. \(t=2\).
Gain smoothing results in a decrease in coverage time.

\textbf{Gradient Averaging}

Due to its stochastic nature and noisy measurements, the gradients
\(\hat{g}(\hat{w})\) can be approximately wrongly and hence distort the
convergence direction in SPSA-FS algorithm. To mitigate such side
effect, the current and the previous \(m\) gradients are averaged as a
gradient estimate at the current iteration:

\begin{center}
  \begin{align}
    \hat{g_k}(\hat{w_k}) = \frac{\sum_{n=k-m}^k{\hat{g_{n}}(\hat{w_{k}})}}{m+1} \label{eqn3.11}
  \end{align}
\end{center}

SPSA-FS is developed to converge much more faster than BSPSA at an small
incremental in the loss function. Algorithm \ref{pseudoCodes2}
summarises the pseudo code for the SPSA-FS Algorithm, which is modified
based on the BSPSA Algorithm (see Algorithm \ref{pseudoCodes}). Note
that in Algorithm \ref{pseudoCodes2}, Steps 13 and 14 correspond to
Equation \ref{eqn3.9} whereas Step 15 correspond to Equation
\ref{eqn3.10}.

\begin{algorithm}
\caption{SPSA-FS Algorithm} \label{pseudoCodes2}
\begin{algorithmic}[1]
\Procedure{\underline{SPSA-FS}($\hat{w}_0$, $c$, $M$)}{}
\BState Initialise $k = 0$, $m=0$
\BState \textbf{do}:
\State Simulate $\Delta_{k, j} \sim \text{Bernoulli}(-1, +1)$ with $\mathbb{P}(\Delta_{k, j}=1) = \mathbb{P}(\Delta_{k, j}=-1) = 0.5$ for $j=1,..p$
\State $\hat{w}^{\pm}_k = \hat{w}_{k} \pm c \Delta_k$
\State $\hat{w}^{\pm}_k = B(\hat{w}^{\pm}_k)$ \Comment{$B( \bullet)$ = component-wise $[0,1]$ operator }
\State $\hat{w}^{\pm}_k = R(\hat{w}^{\pm}_k)$ \Comment{$R( \bullet)$ = component-wise rounding operator}
\State $y^{\pm}_k =\mathcal{L}(\hat{w}_k \pm c_k \Delta_k) \pm \varepsilon_{k}^{\pm}$
\State $\hat{g}_k(\hat{w}_k) =\bigg( \frac{y^{+}_k-y^{-}_k}{2c}\bigg)[\Delta_{k1}^{-1},...,\Delta_{kp}^{-1}]^{T}$ \Comment{$\hat{g}_k(\hat{w}_k)$ = the gradient estimate}

\State $\hat{g_k}(\hat{w_k}) = \frac{1}{m+1}\sum_{n=k-m}^k{\hat{g_{n}}(\hat{w_{k}})}$ \Comment{Gradient Averaging}

\State $\hat{a}_k = \frac{\nabla{\hat{w}}^T\nabla{\hat{g}(\hat{w})}}{\nabla{\hat{g}^T(\hat{w})}\nabla{\hat{g}(\hat{w})}}$ \Comment{$\hat{a}_k$ = BB Step Size}

\If{$\hat{a}_{k}<0$}
  \State $\hat{a}_k = \max \bigg(\min{\{\hat{a}_{k}\}},\min\{\hat{a}_k, \max{\{\hat{a}_{k}\}}\}\bigg)$ 
\EndIf

\State $\hat{a}_k = \frac{1}{t+1}\sum_{n=k-t}^k{\hat{a}_{n}}$ for $t = \min\{2, k\}$ \Comment{Gain Smoothing}

\State $\hat{w}^{\pm}_k = \hat{w}_{k} \pm  a_k \hat{g}_k(\hat{w}_k)$

\State $k = k + 1$, $m = k$
\BState \textbf{while} ($k < M$)
\BState \textbf{Output}: $\hat{w}^{\pm}_M = R(\hat{w}^{\pm}_M)$
\EndProcedure
\end{algorithmic}
\end{algorithm}

\section{SPSA-FS vs BSPSA Algorithms}\label{spsa-fs-vs-bspsa-algorithms}

\begin{figure}

{\centering \includegraphics{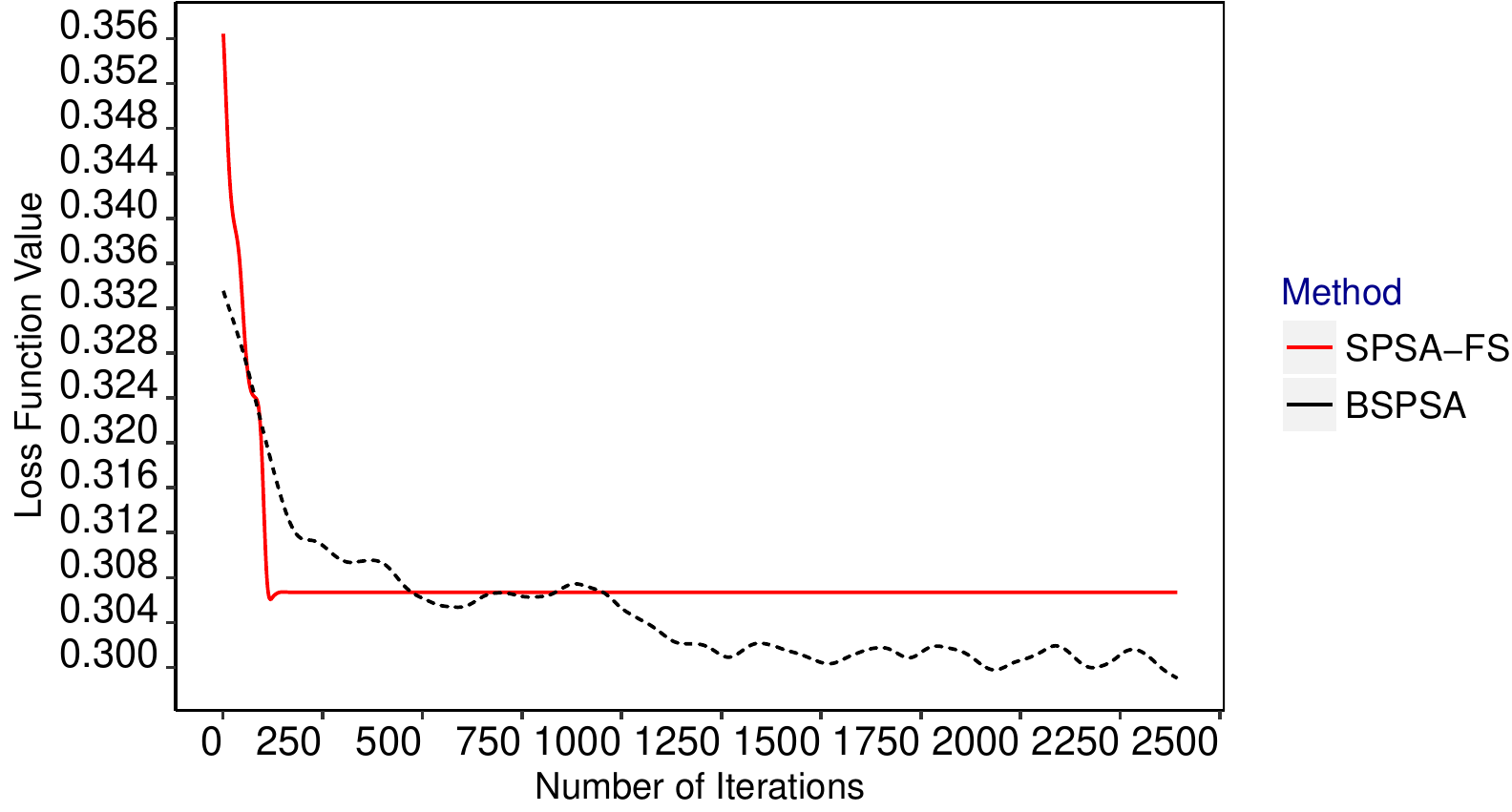} 

}

\caption{\label{figA}Covergence Time Comparison between SPSA-FS and BSPSA on Arrhythmia dataset. Using decision-tree, SPSA-FS hit its lowest loss function value before reaching 250 iterations. BSPSA required around 500 iterations to achieve the same loss function value although it outperformed the SPSA-FS algorithm if more iterations were allowed.}\label{fig:figA}
\end{figure}

SPSA-FS can locate a solution around 400\% faster than BSPSA by losing
only 2\% in the prediction accuracy given the same dataset. In other
words, the SPSA-FS algorithm is five times faster than the BSPSA
algorithm to reach the same loss function value or the accuracy rate. In
practice, the difference might rachet up to 20 times with a minimal drop
in the accuracy around 2 \%. For illustration, we experimented with the
decision (or recursive partitioning) tree as a wrapper on Arrhythmia
dataset provided by Guvenir et al.
(\protect\hyperlink{ref-guvenir1997supervised}{1997}) accessible at
\href{http://archive.ics.uci.edu/ml}{UCI Machine Learning Repository}
(Lichman \protect\hyperlink{ref-UCI}{2013}). This dataset contains 279
features. Figure \ref{figA} compares the performance of two algorithms
by evaluating their inaccuracy rates (loss function values) at each
iteration. BSPSA found the lowest and hence the best loss function but
five times slower than SPSA-FS Algorithm. It was 20 times slower with
regard to overall calculation period.

\newpage

\chapter{Wrapper Comparison}\label{section4}

Supervised learning problems can fall into two broad categories:
classification and regression. The target or dependent feature is a
binary, nominal or ordinal variable in a classification task while it is
a continuous variable in a regression problem. Apart from feature
selection, classification problems have another important aspect:
feature ranking. While feature selection aims to determine the optimal
subset of predictive features, feature ranking measures how important
each feature from the specified set explains the target feature. For
comparability, we divided the wrapper comparison experiments into three
sections below:

\begin{itemize}
\tightlist
\item
  \protect\hyperlink{feature-selection-in-classification-problems}{Feature
  Selection in Classification Problems}
\item
  \protect\hyperlink{feature-ranking-in-classification-problems}{Feature
  Ranking in Classification Problems}
\item
  \protect\hyperlink{feature-selection-in-regression-problems}{Feature
  Selection in Regression Problems}
\end{itemize}

We ran the wrapper comparison experiments using the open dataset
accessible from the following sources:

\begin{itemize}
\tightlist
\item
  \href{http://archive.ics.uci.edu/ml}{UCI Machine Learning Repository}
  (Lichman \protect\hyperlink{ref-UCI}{2013})
\item
  \href{http://www.dcc.fc.up.pt/~ltorgo/Regression/DataSets.html}{DCC
  Regression DataSets} (Torgo \protect\hyperlink{ref-DCC}{2017})
\item
  \href{http://featureselection.asu.edu/index.php}{Scikit-feature
  feature selection repository at ASU} (J. Li et al.
  \protect\hyperlink{ref-ASU}{2016})
\end{itemize}

\hypertarget{feature-selection-in-classification-problems}{\section{Feature
Selection in Classification
Problems}\label{feature-selection-in-classification-problems}}

We selected nine (9) datasets for feature selection (see Table
\ref{tabA}). For each dataset, we implemented four classifiers namely
Recursive Partitioning for Classification (R.Part), K-Nearest Neighbours
(KNN), Naïve Bayes (NB), and Support Vector Machine (SVM). For each
classifier, we considered three main wrapper methods:

\begin{itemize}
\tightlist
\item
  SPSA;
\item
  SFS: Sequential feature selection; and
\item
  Full as the baseline benchmark.
\end{itemize}

For some datasets, we compared four following additional wrappers below:

\begin{itemize}
\tightlist
\item
  GA: Genetic Algorithm;
\item
  SBS: Sequential Backward Selection;
\item
  SFFS: Sequential Feature Forward selection; and
\item
  SFBS: Sequential Feature Backward selection.
\end{itemize}

Besides the mean classification error rate, we also considered the mean
runtime of the learning process to assess feature selection performance.
Despite the mixed results, SPSA-FS managed to balance accuracy and
runtime on average. In the scenario where SPSA-FS outperformed in the
accuracy, it required slightly more or approximately runtime of other
wrappers. When SPSA-FS trailed behind other wrappers in term of
accuracy, it did not cost too much runtime. Such empirical results were
consistent with the theoretical design of SPSA-FS where the BB method
helps reduce the computational cost by sacrificing a minimal amount of
accuracy.

\begin{table}
    \centering
    \begin{tabular}{ | l | l | l | l | l |}
    \hline
      dataset & $p$ & $N$ & Source & Figure \\
      \hline
Arrhythmia & 279 & 452 & \href{https://archive.ics.uci.edu/ml/datasets/arrhythmia}{UCI} & Figure \ref{FRfig1} \\
Glass & 9 & 214 & \href{https://archive.ics.uci.edu/ml/datasets/glass+identification}{UCI} & Figure \ref{FRfig2} \\
Heart & 13 & 270 & \href{http://archive.ics.uci.edu/ml/datasets/statlog+(heart)}{UCI} & Figure \ref{FRfig3} \\
Ionosphere & 34 & 351 & \href{https://archive.ics.uci.edu/ml/datasets/ionosphere}{UCI} & Figure \ref{FRfig4} \\
Libras & 90 & 360 & \href{https://archive.ics.uci.edu/ml/datasets/Libras+Movement}{UCI} & Figure \ref{FRfig5} \\
Musk (Version 1) & 166 & 476 & \href{https://archive.ics.uci.edu/ml/machine-learning-databases/musk/ }{UCI} & Figure \ref{FRfig6} \\
Sonar & 60 & 208 & \href{http://archive.ics.uci.edu/ml/datasets/connectionist+bench+(sonar,+mines+vs.+rocks)}{UCI} & Figure \ref{FRfig7} \\
Spam Base & 57 & 4601 & \href{https://archive.ics.uci.edu/ml/datasets/spambase}{UCI} & Figure \ref{FRfig8} \\
Vehicle & 18 & 946 & \href{https://archive.ics.uci.edu/ml/datasets/Statlog+(Vehicle+Silhouettes)}{UCI} & Figure \ref{FRfig9} \\
      \hline
    \end{tabular}
    \caption{Feature Selection Classification datasets. $p$ represents the number of explanatory feature excluding exclude the response variables and identifier attributes; $N$ denotes the number of observations.}
    \label{tabA}
\end{table}

\begin{figure}
\centering
\includegraphics{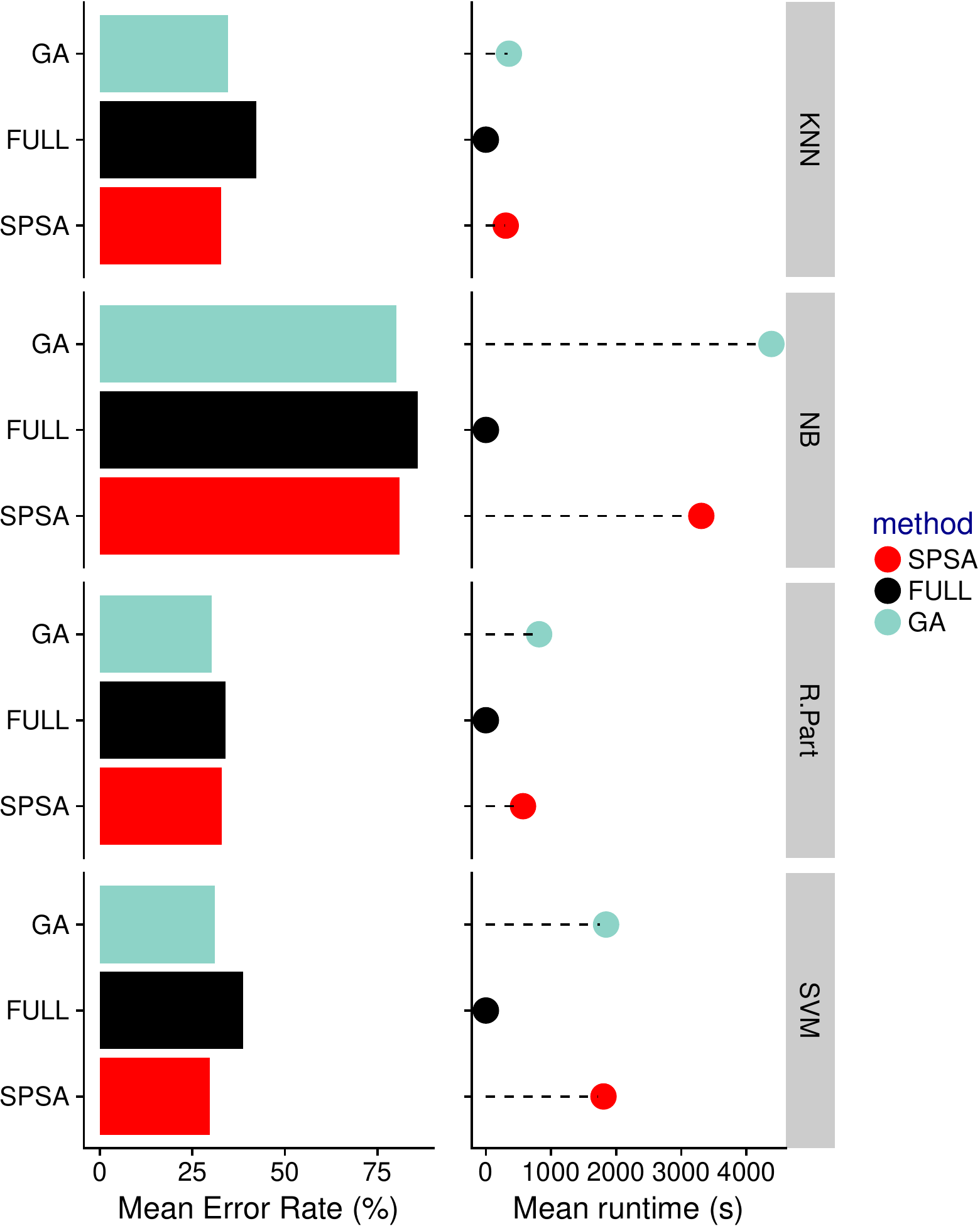}
\caption{\label{FRfig1}Feature Selection Performance Result on
Arrhythmia}
\end{figure}

\begin{figure}
\centering
\includegraphics{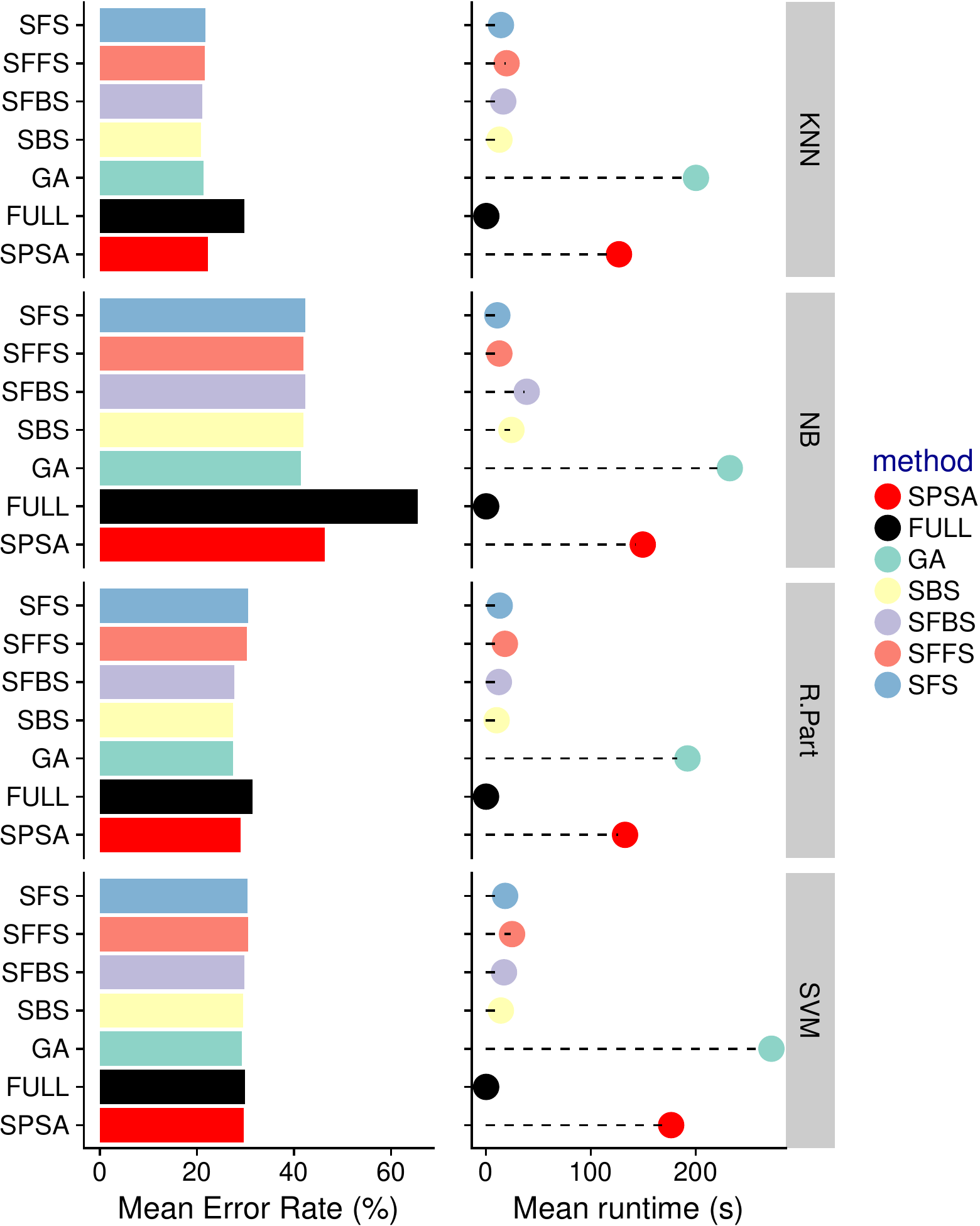}
\caption{\label{FRfig2}Feature Selection Performance Result on Glass}
\end{figure}

\begin{figure}
\centering
\includegraphics{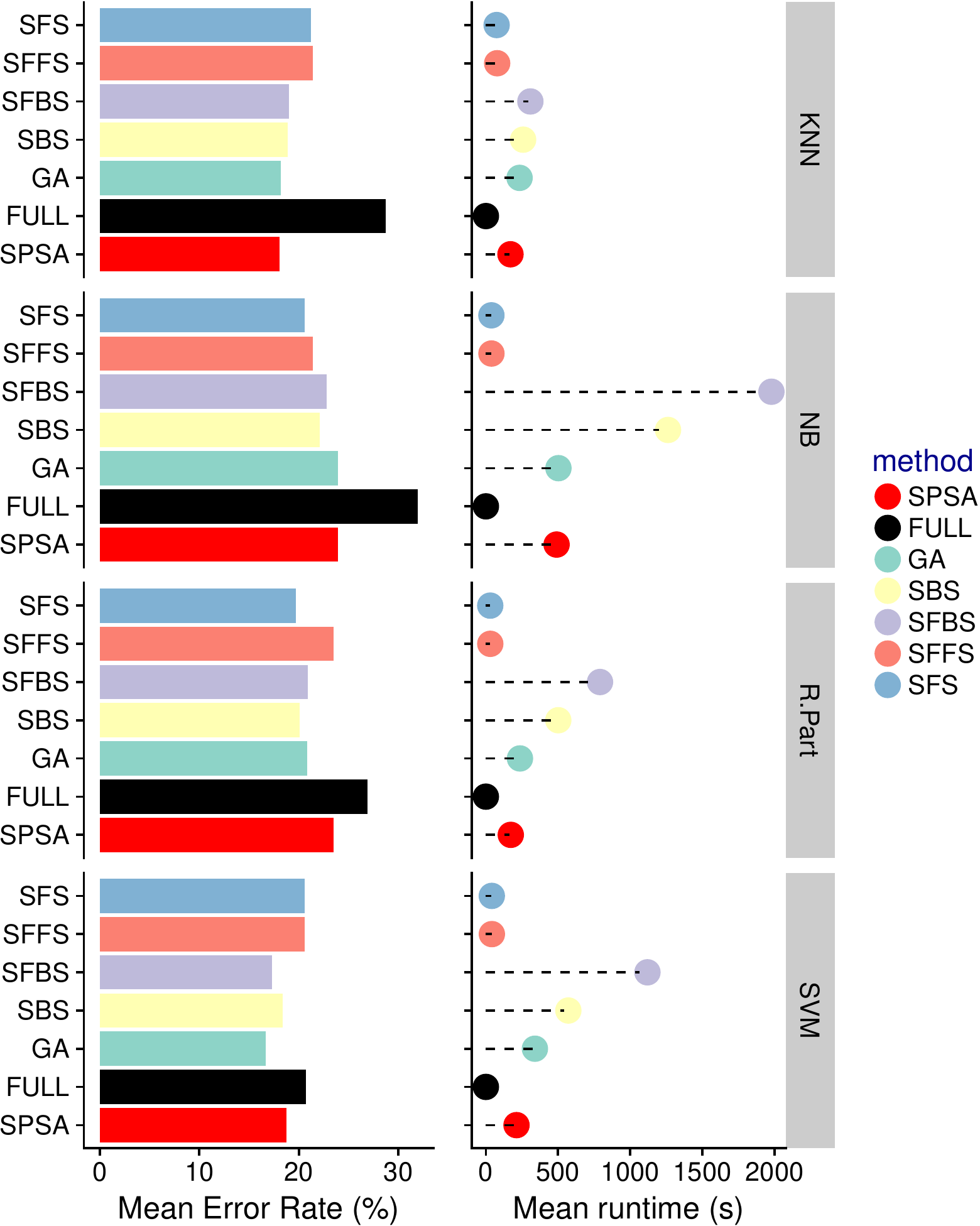}
\caption{\label{FRfig3}Feature Selection Performance Result on Heart}
\end{figure}

\begin{figure}
\centering
\includegraphics{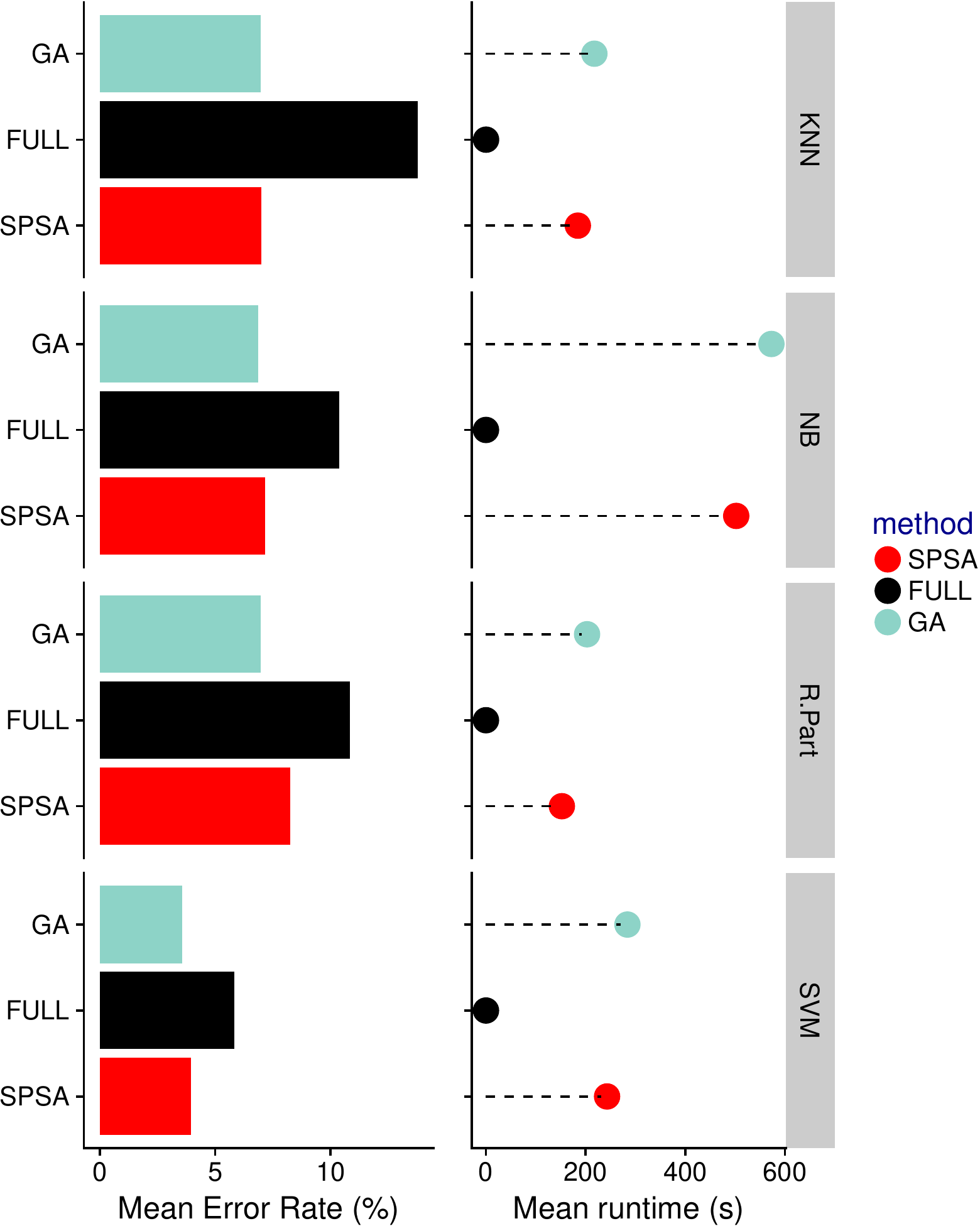}
\caption{\label{FRfig4}Feature Selection Performance Result on
Ionospehere}
\end{figure}

\begin{figure}
\centering
\includegraphics{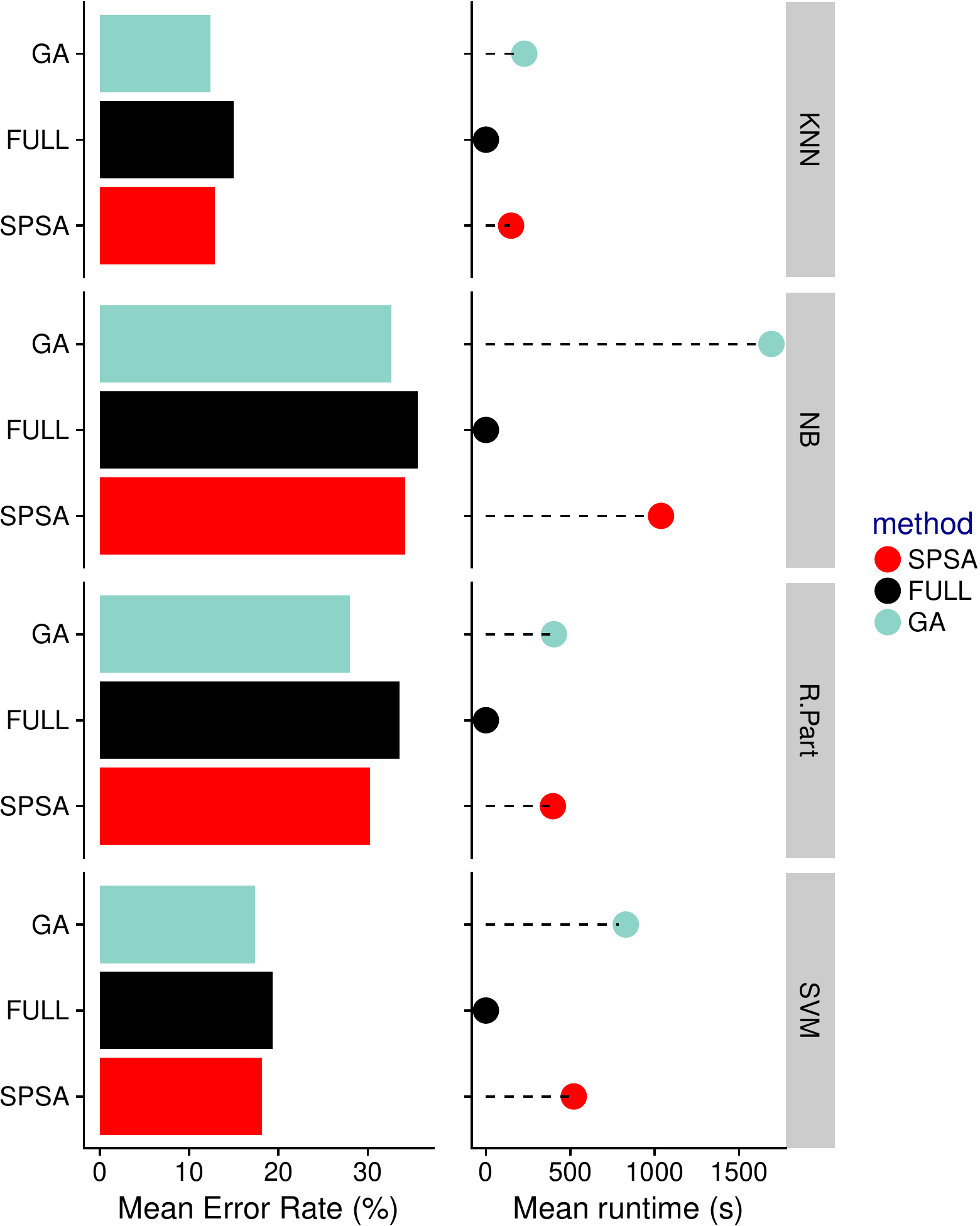}
\caption{\label{FRfig5}Feature Selection Performance Result on Libras}
\end{figure}

\begin{figure}
\centering
\includegraphics{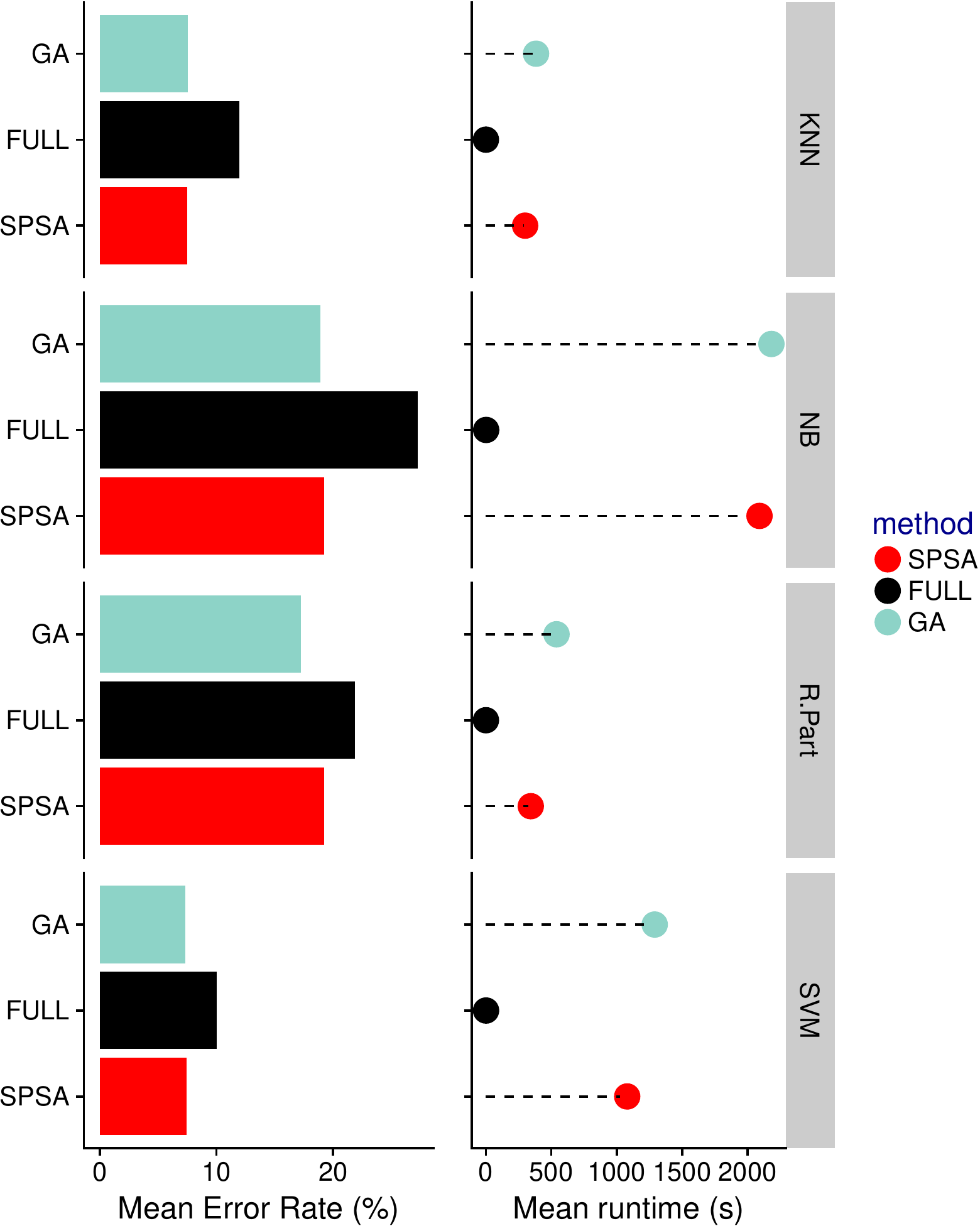}
\caption{\label{FRfig6}Feature Selection Performance Result on Musk}
\end{figure}

\begin{figure}
\centering
\includegraphics{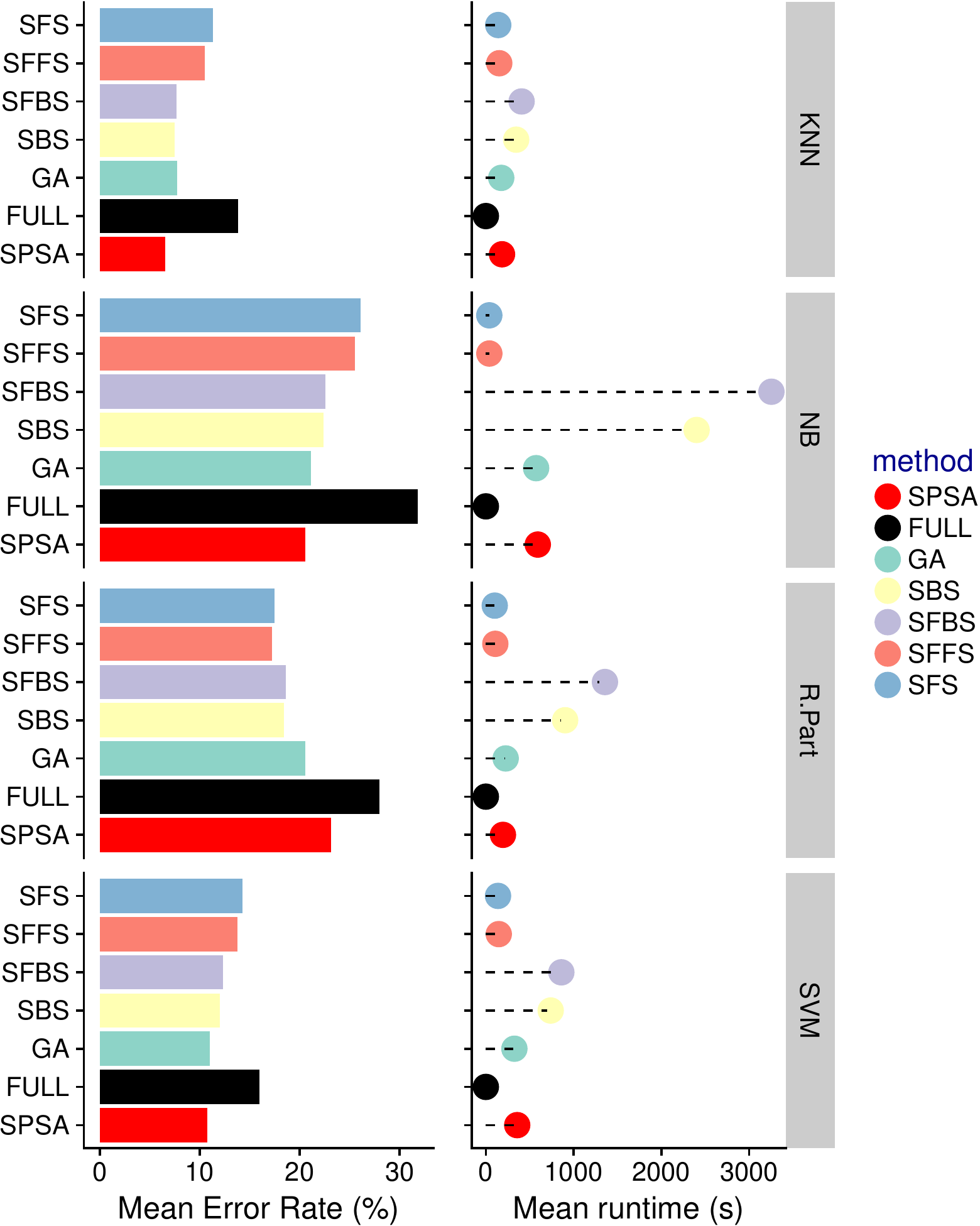}
\caption{\label{FRfig7}Feature Selection Performance Result on Sonar}
\end{figure}

\begin{figure}
\centering
\includegraphics{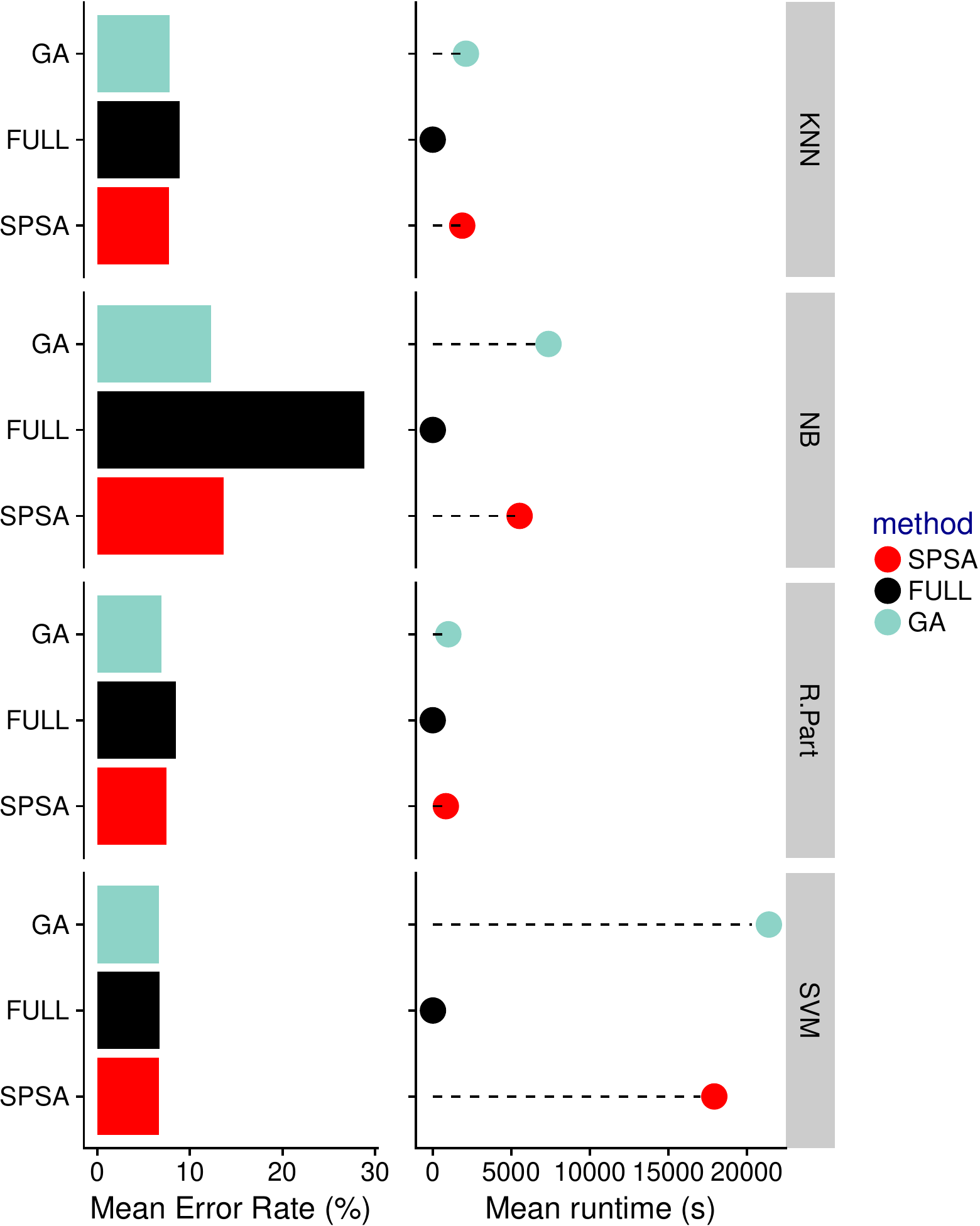}
\caption{\label{FRfig8}Feature Selection Performance Result on Spam
Base}
\end{figure}

\begin{figure}
\centering
\includegraphics{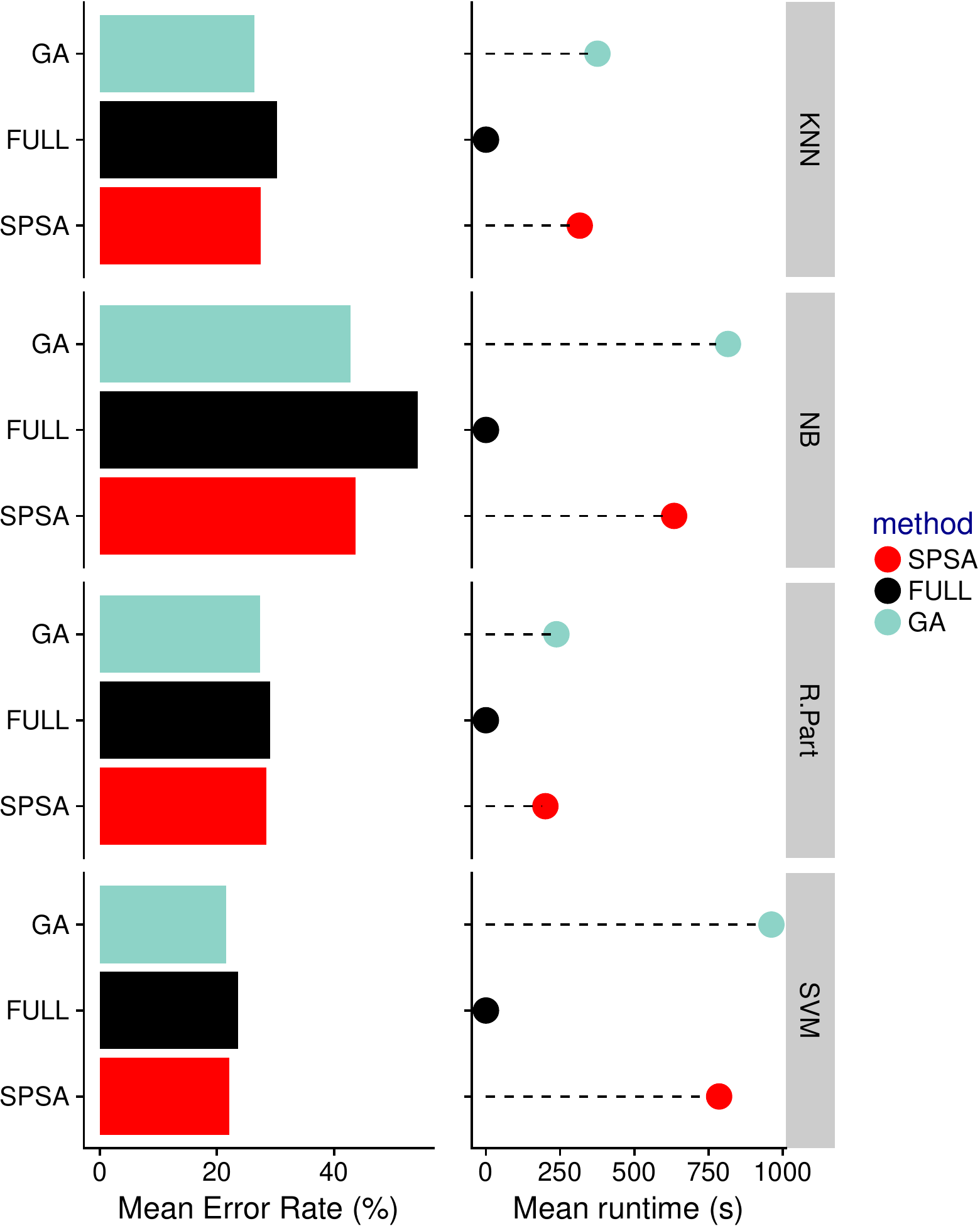}
\caption{\label{FRfig9}Feature Selection Performance Result on Vehicle}
\end{figure}

\hypertarget{feature-ranking-in-classification-problems}{\section{Feature
Ranking in Classification
Problems}\label{feature-ranking-in-classification-problems}}

Table \ref{tabB} delineates eight (8) datasets for feature ranking. For
each dataset, we applied four (4) classifiers which were Decision Tree
(DT), K-Nearest Neighbours (KNN), Naïve Bayes (NB), and Support Vector
Machine (SVM). For each classifier, using the mean classification error
rate, we compared five wrapper methods below:

\begin{itemize}
\tightlist
\item
  SPFS: SPSA as Feature Selection;
\item
  RFI: Random Forest Importance;
\item
  Chi.Sq: Chi-Squared;
\item
  Info.Gain: Information Gain; and
\item
  Full as the baseline benchmark.
\end{itemize}

As shown in Table \ref{tabB}, each dataset has a large number of
features, especially Orl and AR10p which suffer the high dimensionality
curse, i.e. \(p > N\). To illustrate the difference in feature ranking
by wrappers, we capped the number of features used say \(m\) in each
classifier and utilised each wrapper to return the top \(m\) important
features. For example, consider Sonar dataset which consists of 60
features and \(m=5\). Each wrapper would rank the top 5 important
features out of 60 which yield the lowest accuracy rate in classifying
the type of rock from Sonar data.

\begin{figure}
\centering
\includegraphics{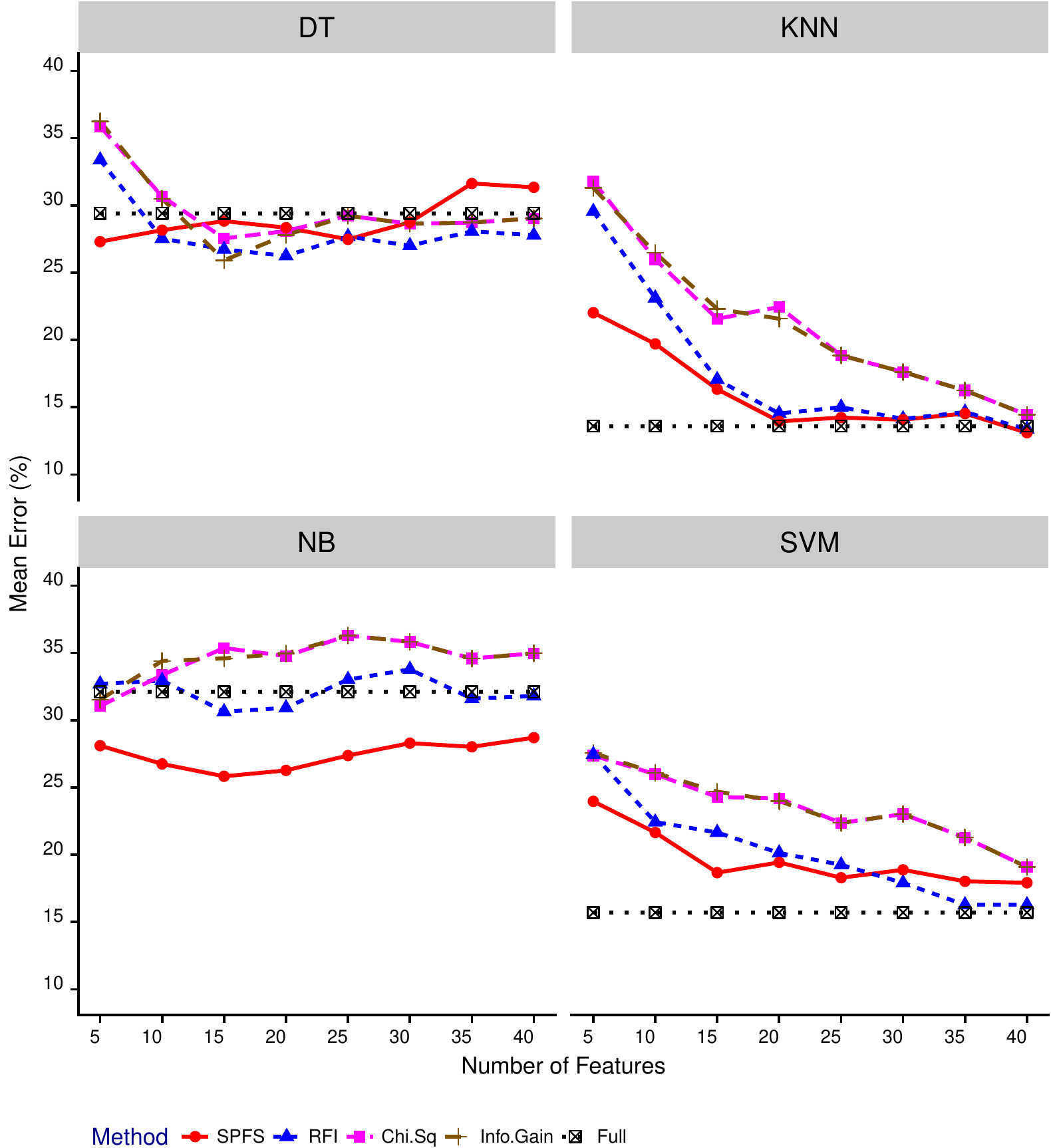}
\caption{\label{fig9}Wrapper Misclassification Error on Sonar dataset by
Types of Classifiers}
\end{figure}

For completeness, we ran the experiment on a series of \(m\) starting
from 5 up to 40 features in an increment of 5. That is,
\(m = \{5, 10, 15, ... ,40\}\). We also compared the wrapper performance
to the baseline benchmark, which incorporated all features. Unlike
\(R^{2}\) used in regression problems, additional explanatory features
do not necessarily improve the accuracy rate. For example, as depicted
in Figure \ref{fig9}, NB classifier committed more misclassifications on
Sonar data when more than 30 features were used in each wrapper.

\begin{figure}
\centering
\includegraphics{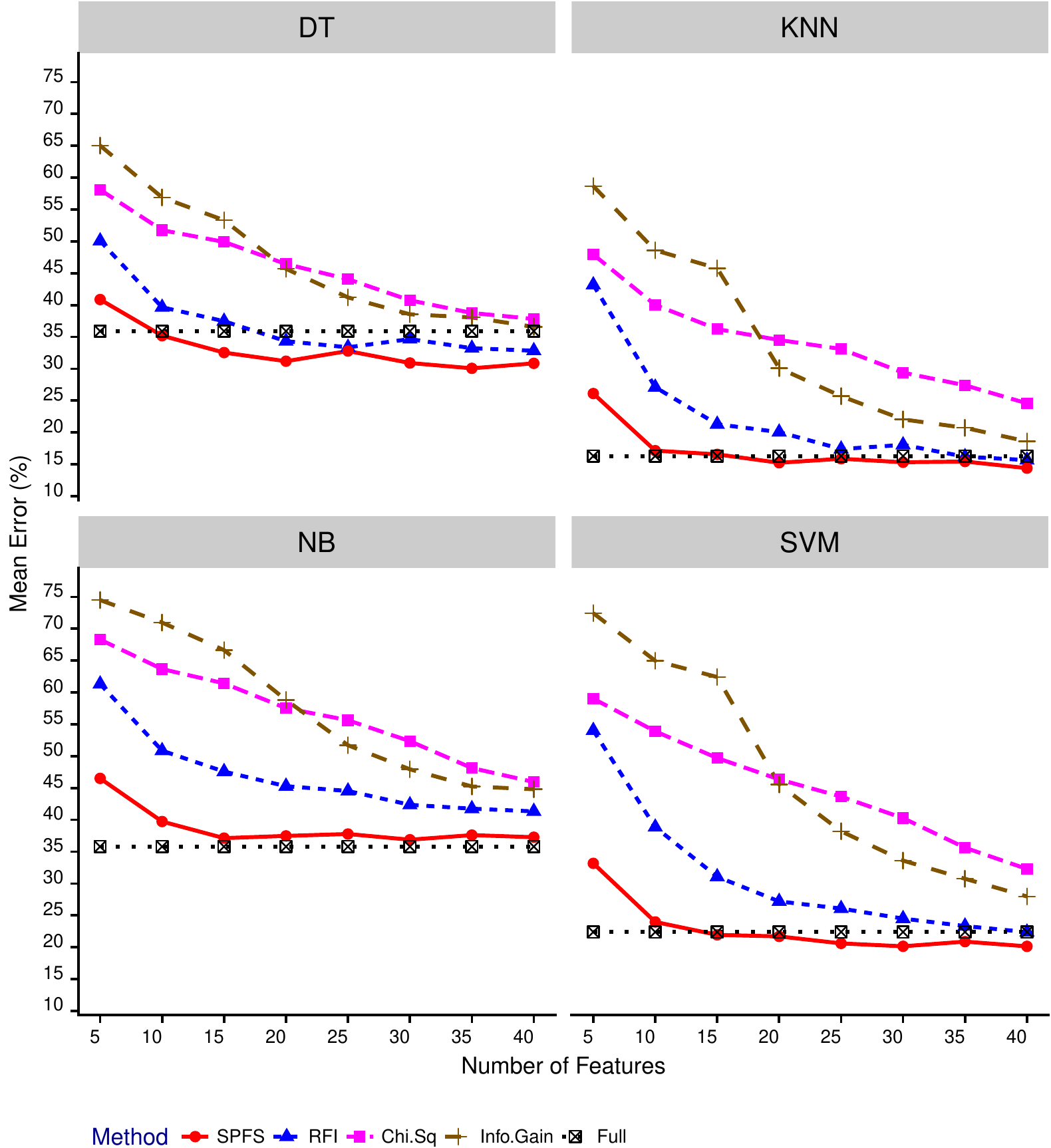}
\caption{\label{fig10}Wrapper Misclassification Error on Libras dataset
by Types of Classifiers}
\end{figure}

\begin{figure}
\centering
\includegraphics{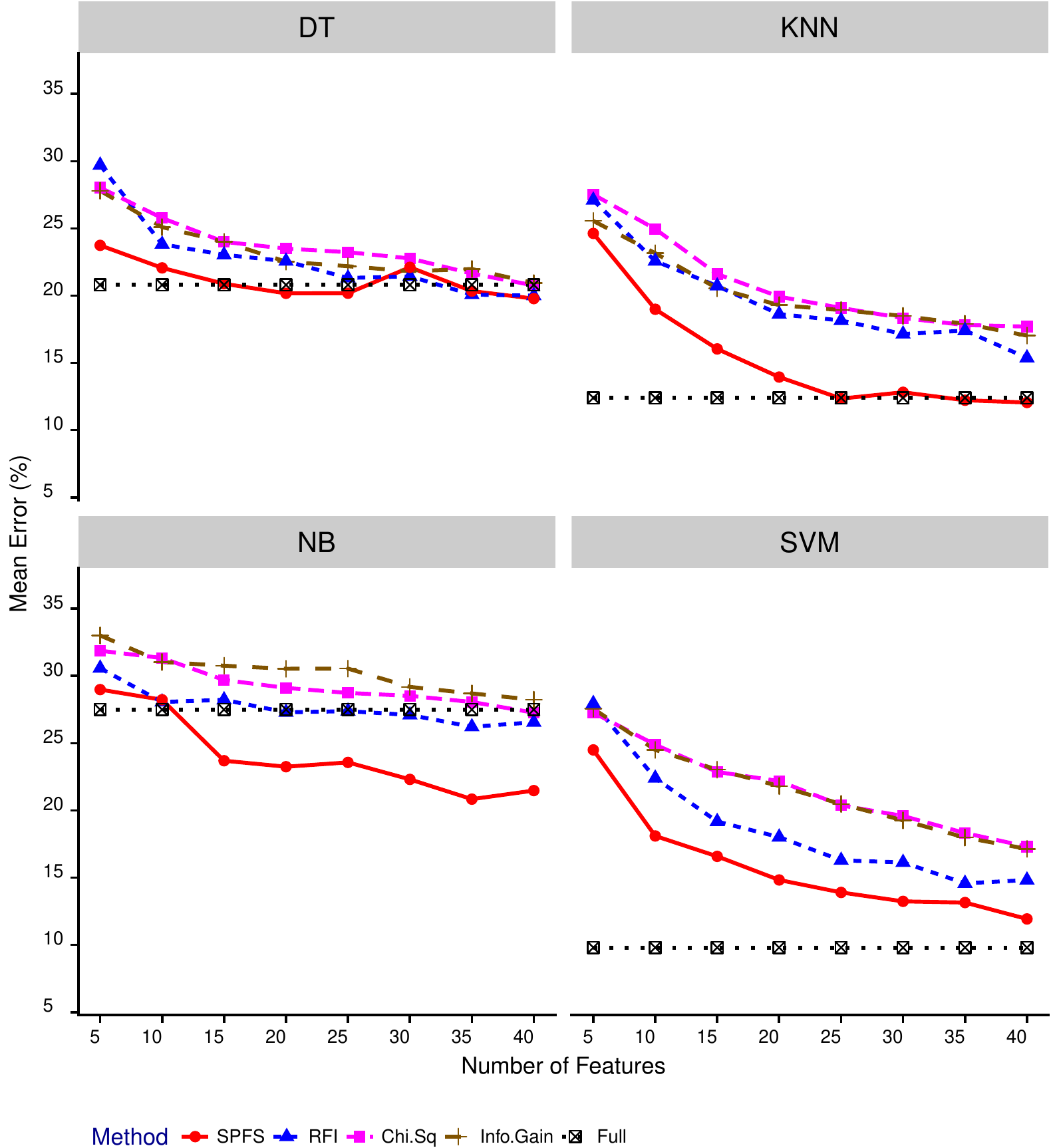}
\caption{\label{fig11}Wrapper Misclassification Error on Musk dataset by
Types of Classifiers}
\end{figure}

\begin{figure}
\centering
\includegraphics{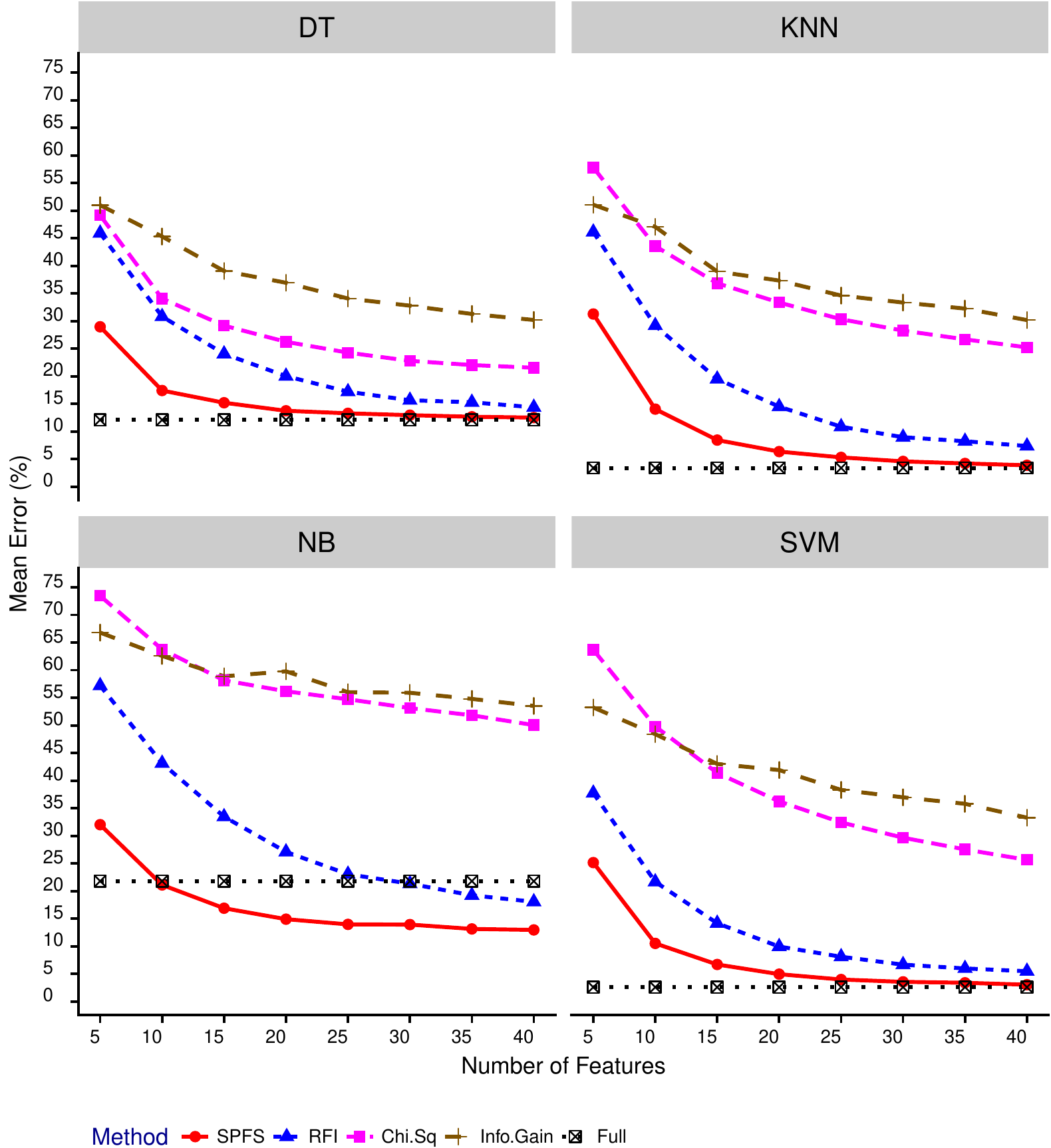}
\caption{\label{fig12}Wrapper Misclassification Error on USPS dataset by
Types of Classifiers}
\end{figure}

\begin{figure}
\centering
\includegraphics{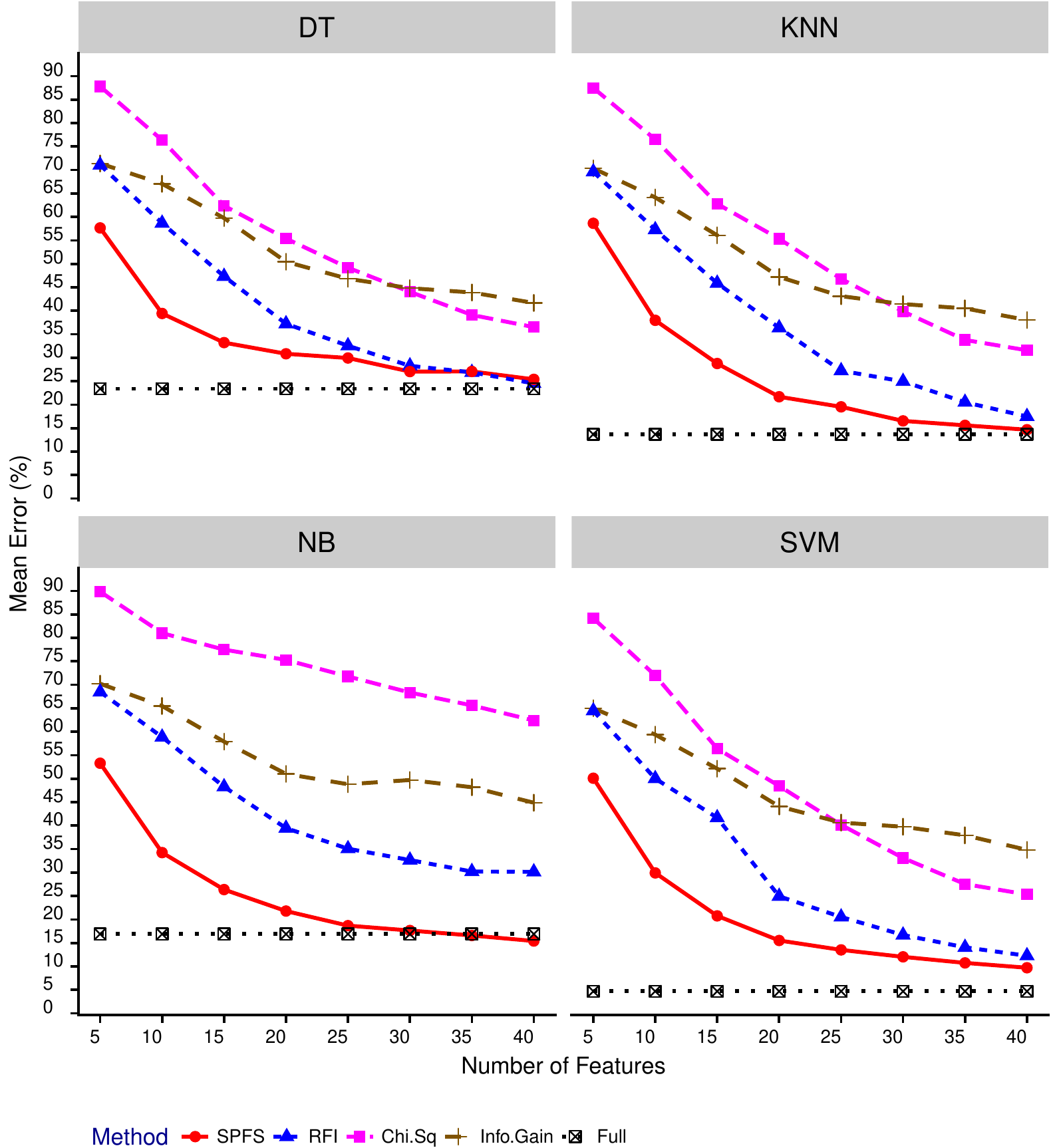}
\caption{\label{fig13}Wrapper Misclassification Error on Isolet dataset
by Types of Classifiers}
\end{figure}

\begin{figure}
\centering
\includegraphics{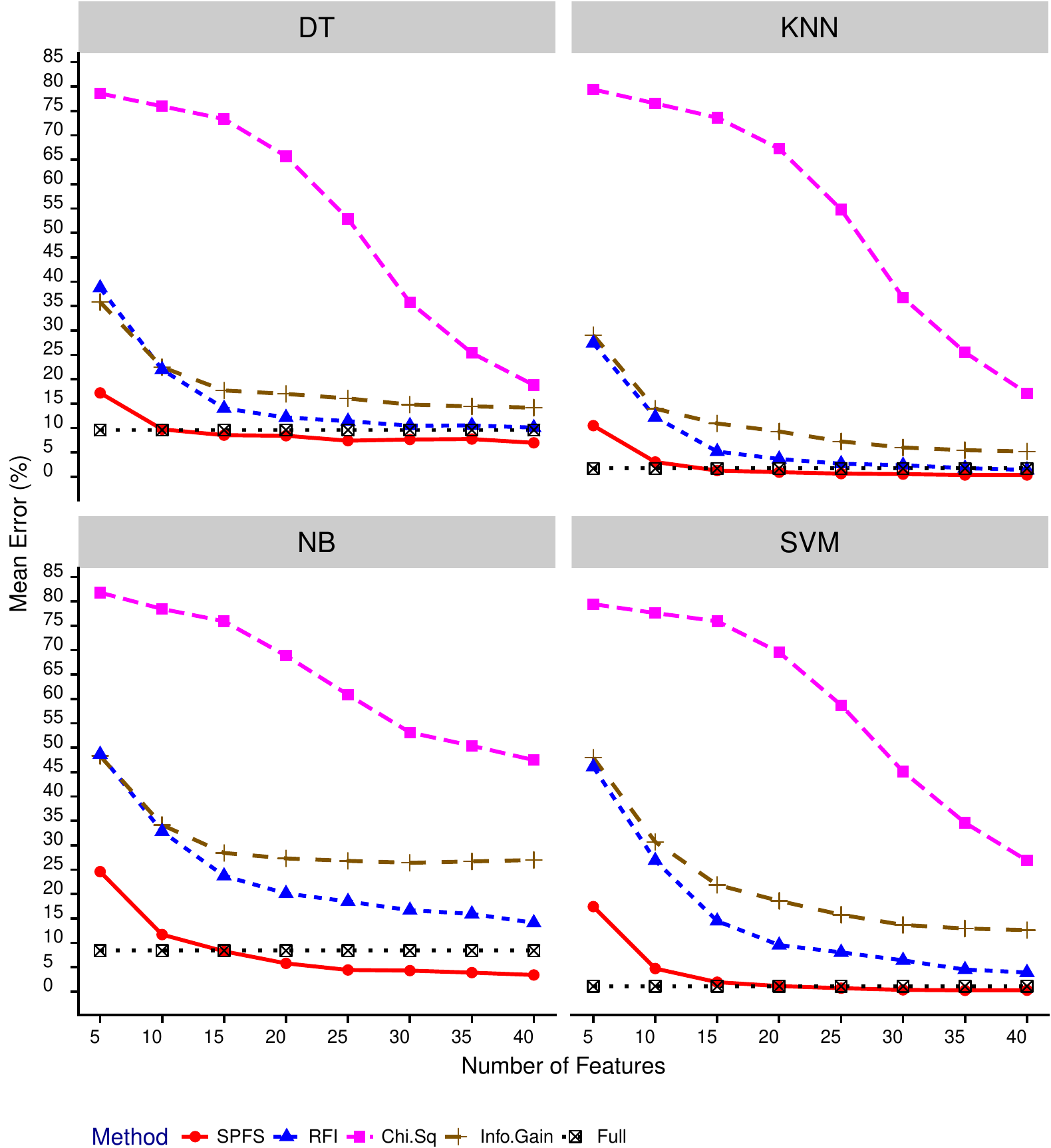}
\caption{\label{fig14}Wrapper Misclassification Error on Coil20 dataset
by Types of Classifiers}
\end{figure}

\begin{figure}
\centering
\includegraphics{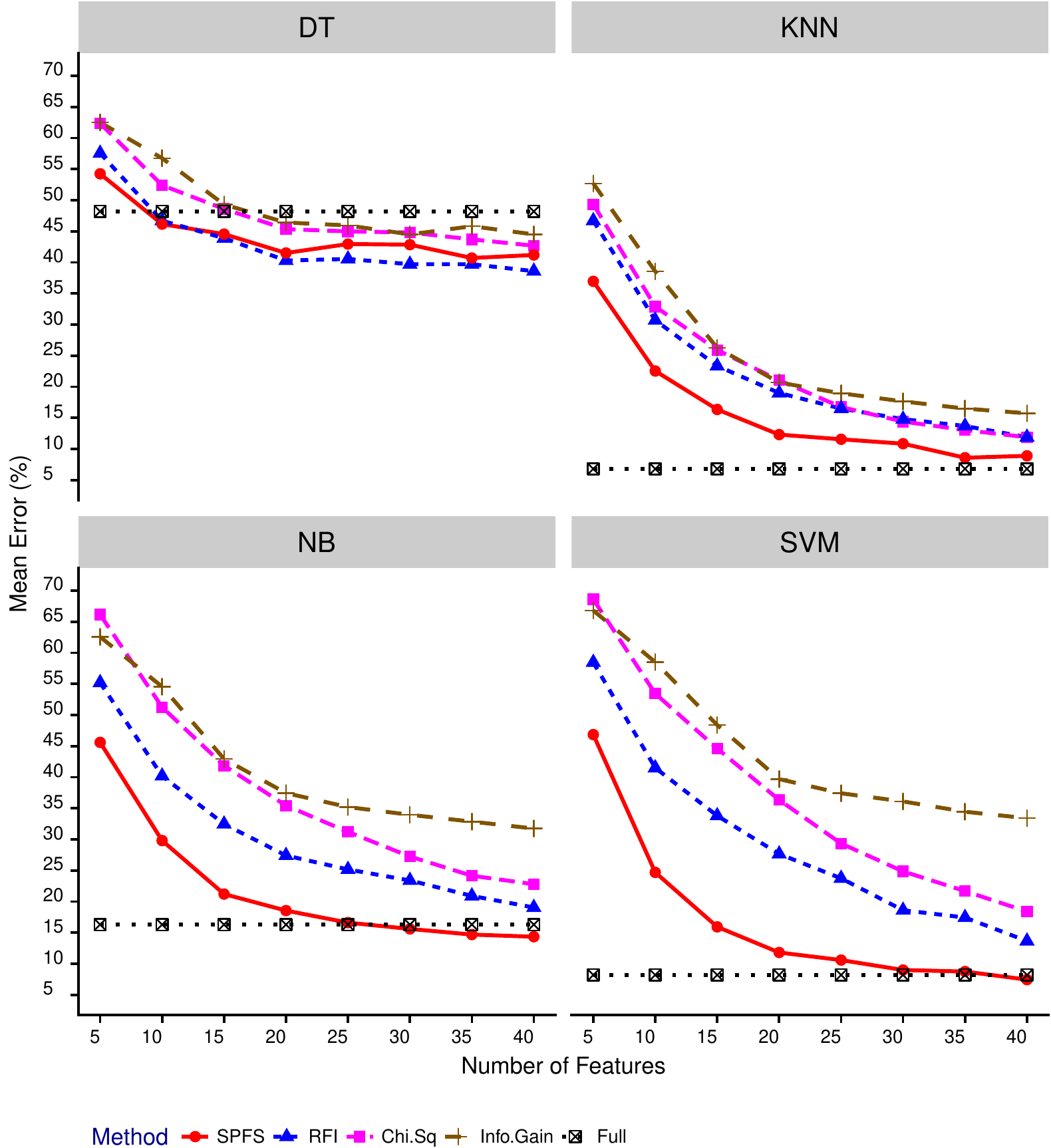}
\caption{\label{fig15}Wrapper Misclassification Error on Orl dataset by
Types of Classifiers}
\end{figure}

\begin{figure}
\centering
\includegraphics{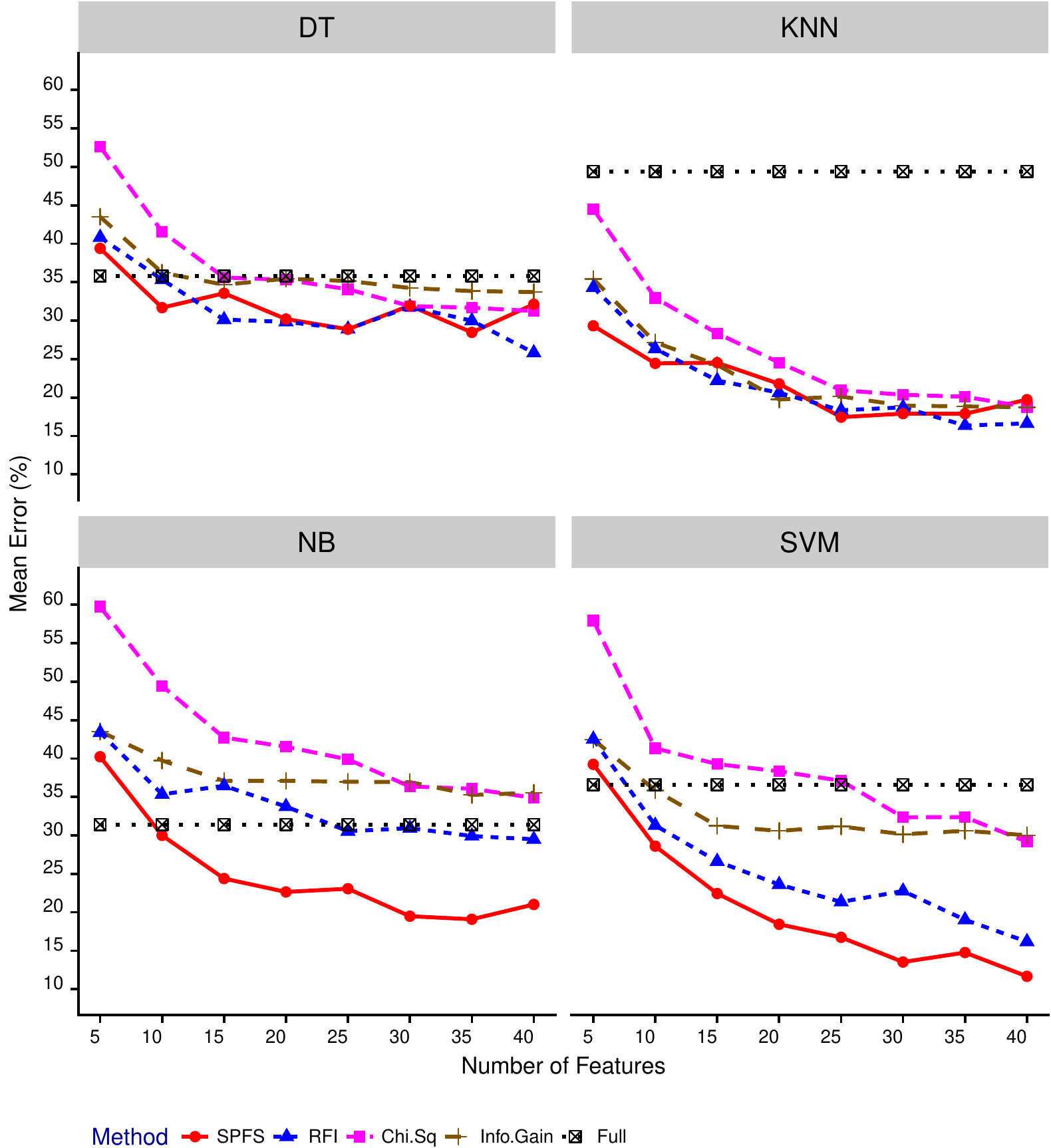}
\caption{\label{fig16}Wrapper Misclassification Error on AR10p dataset
by Types of Classifiers}
\end{figure}

From Figures \ref{fig9} to \ref{fig16}, we inferred that:

\begin{enumerate}
\def\labelenumi{\arabic{enumi}.}
\tightlist
\item
  With some exceptions, the accuracy rates tended to decrease as the
  number of features increased albeit at a lower rate.
\item
  SPSA-FS outperformed other wrapper methods in most data sets but did
  not consistently beat the baseline due to the choice of classifier.
\end{enumerate}

\begin{table}
    \centering
    \begin{tabular}{ | l | l | l | l | l |}
    \hline
      Dataset & $p$ & $N$ & Source & Figure \\
      \hline
        Sonar & 60 & 208 & \href{http://archive.ics.uci.edu/ml/datasets/connectionist+bench+(sonar,+mines+vs.+rocks)}{UCI} & Figure \ref{fig9} \\
        Libras & 90 & 360 & \href{https://archive.ics.uci.edu/ml/datasets/Libras+Movement}{UCI} & Figure \ref{fig10} \\
        Musk (Version 1) & 166 & 476 & \href{https://archive.ics.uci.edu/ml/machine-learning-databases/musk/}{UCI} & Figure \ref{fig11} \\
        Usps & 256 &    9298 & \href{http://featureselection.asu.edu/datasets.php}{ASU} & Figure \ref{fig12} \\
        Isolet & 617 & 1560 & \href{http://featureselection.asu.edu/datasets.php}{ASU} & Figure \ref{fig13} \\
        Coil20 & 1024 & 1440 & \href{http://featureselection.asu.edu/datasets.php}{ASU} & Figure \ref{fig14} \\
        Orl & 1024 & 400 & \href{http://featureselection.asu.edu/datasets.php}{ASU} & Figure \ref{fig15} \\
        AR10p & 2400 & 130 & \href{http://featureselection.asu.edu/datasets.php}{ASU} & Figure \ref{fig16} \\
      \hline
    \end{tabular}
    \caption{Feature Ranking Classification datasets. $p$ represents the number of explanatory feature excluding exclude the response variables and identifier attributes; $N$ denotes the number of observations.}
    \label{tabB}
\end{table}

\hypertarget{feature-selection-in-regression-problems}{\section{Feature
Selection in Regression
Problems}\label{feature-selection-in-regression-problems}}

The regression experiment involved eight datasets. Table \ref{tab1}
describes each dataset and their sources. Using a typical linear
regression model, we ran the following feature selection methods for
each dataset:

\begin{itemize}
\tightlist
\item
  SPSA-FS
\item
  Minimum Redundancy Maximum Relevance (mRMR), proposed by C. Ding and
  Peng (\protect\hyperlink{ref-mrmr}{2003})
\item
  RELIEF algorithm, which is first introduced by Kira and Rendell
  (\protect\hyperlink{ref-relief}{1992})
\item
  Linear Correlation
\end{itemize}

\begin{table}
    \centering
    \begin{tabular}{ | l | p{2.5cm} | p{2.5cm} | l | l |}
    \hline
      Dataset & $p$ & $N$ & Source & Figure \\
      \hline
      Ailerons & 39 & 13750 & DCC & Figure \ref{fig1} \\
      CPU ACT & 21 & 8192 & DCC & Figure \ref{fig2} \\
      Elevator & 17 & 16559 & DCC & Figure \ref{fig3} \\
      Boston Housing & 13 & 506 & UCI & Figure \ref{fig4} \\
      Pole Telecomm & 47 & 1500  & DCC & Figure \ref{fig5} \\
      Pyrim & 26 & 74 & DCC & Figure \ref{fig6} \\
      Triazines & 58 & 186 & DCC & Figure \ref{fig7} \\
      Wisconsin Breast Cancer & 32 & 194 & UCI & Figure \ref{fig8} \\
    \hline
    \end{tabular}
    \caption{Regression datasets. $p$ represents the number of explanatory feature excluding exclude the response variables and identifier attributes; $N$ denotes the number of observations.}
    \label{tab1}
\end{table}

For benchmarking, we calculated inaccuracy rate defined by \(1-R^{2}\)
(R-squared) with respect to the number of features used in the
regression. Known as the coefficient of determination, \(R^{2}\)
measures how close the data are to the fitted regression line.
Therefore, a higher \(R^{2}\) implies a lower inaccuracy rate. Note that
\(R^{2}\) can only either increase or remain constant as the number of
explanatory variables or features, \(p\) increases\footnote{We did not
  use the adjusted \(R^{2}\) because it penalises the large number of
  features and hence would defeat our objective of compariing the
  wrapper algorithms.}. For comparative evaluation, we normalised \(p\)
to the percentage of features used since each dataset has a different
number of explanatory features. In all datasets, on average, SPSA-FS
outperformed other wrapper methods even with fewer features. Other
wrapper methods would only catch up with SPSA-FS starting 30 \% of
explanatory features used. At 100 \%, i.e.~when all explanatory features
were used as regressors, all wrapper methods converged to approximately
similar inaccuracy rate due to the nature of \(R^{2}\). The exemplary
performance of SPSA-FS implies it managed to identify the optimal subset
of regressors given the same number of explanatory features compared to
other methods.

\begin{figure}
\centering
\includegraphics{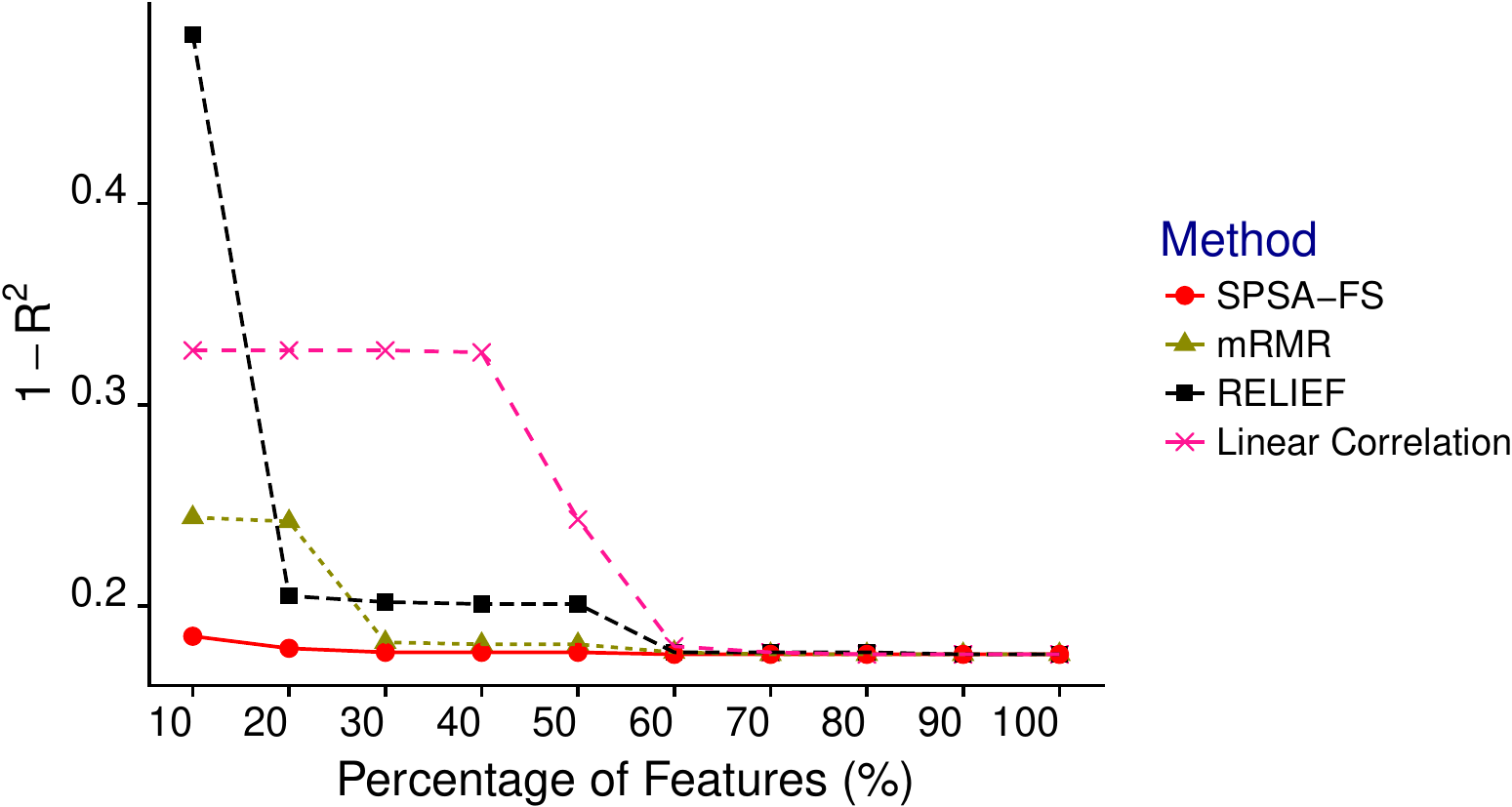}
\caption{\label{fig1}Regression On Ailerons}
\end{figure}

\begin{figure}
\centering
\includegraphics{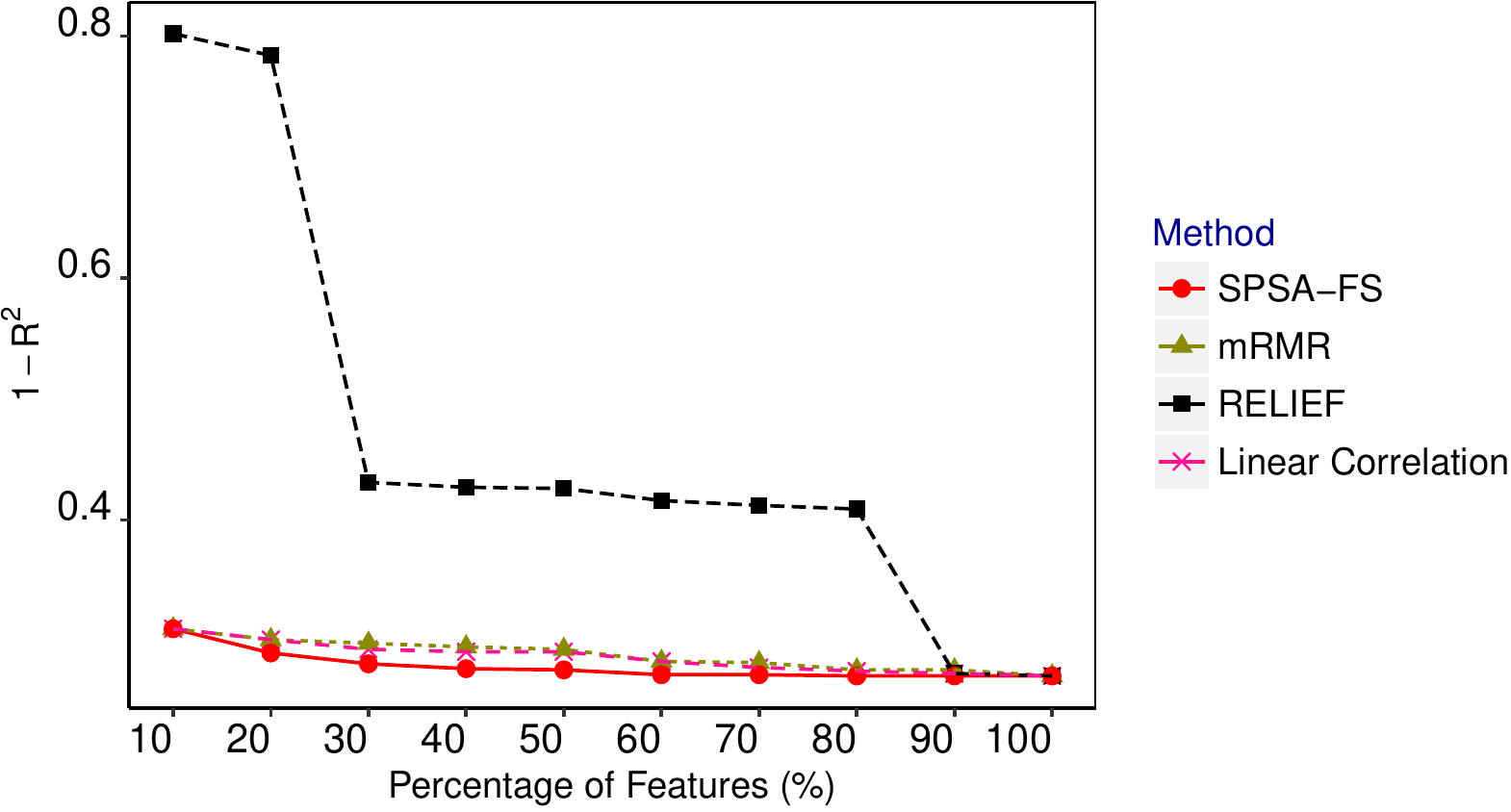}
\caption{\label{fig2}Regression On CPU ACT}
\end{figure}

\begin{figure}
\centering
\includegraphics{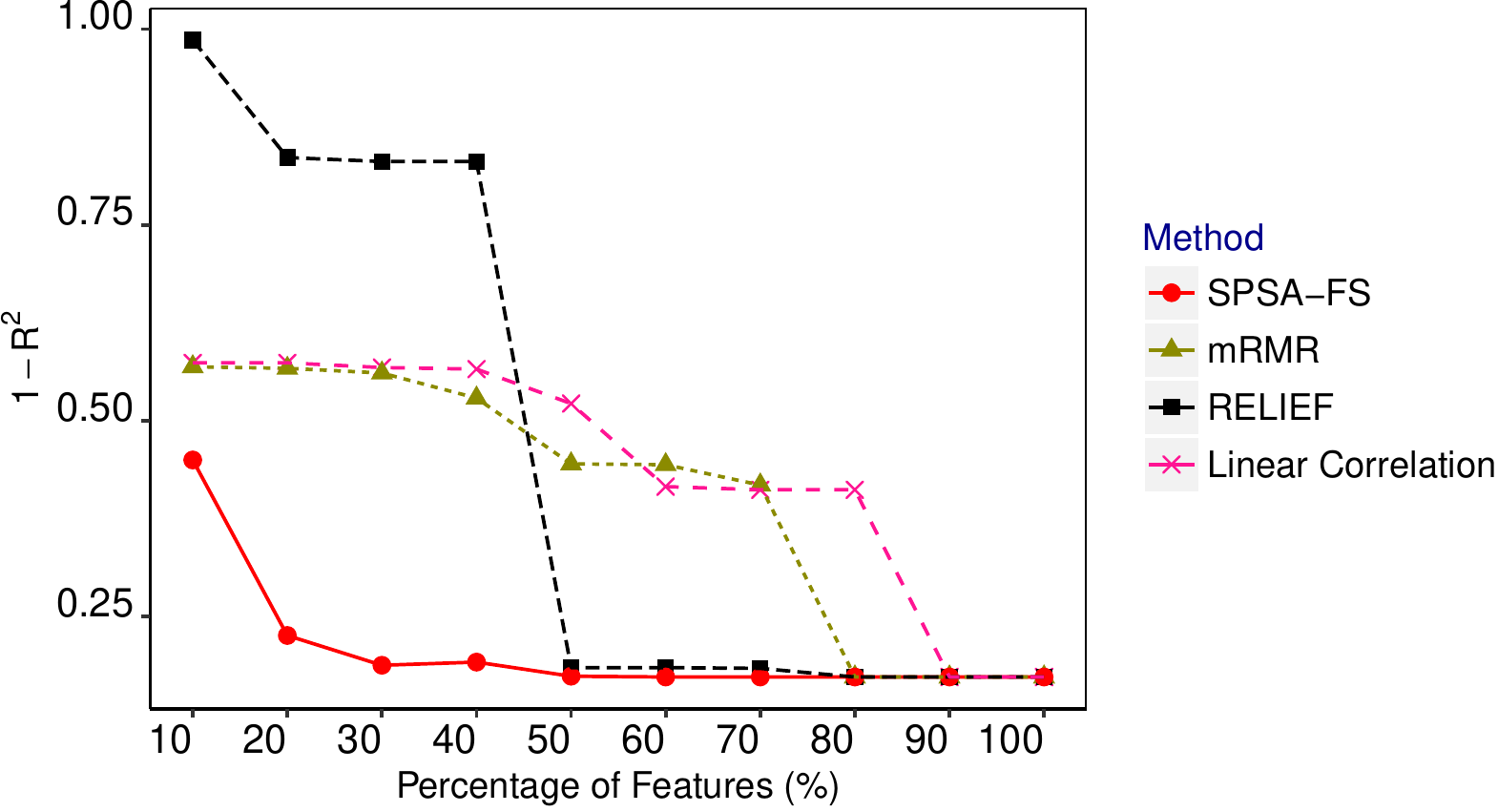}
\caption{\label{fig3}Regression on Elevators}
\end{figure}

\begin{figure}
\centering
\includegraphics{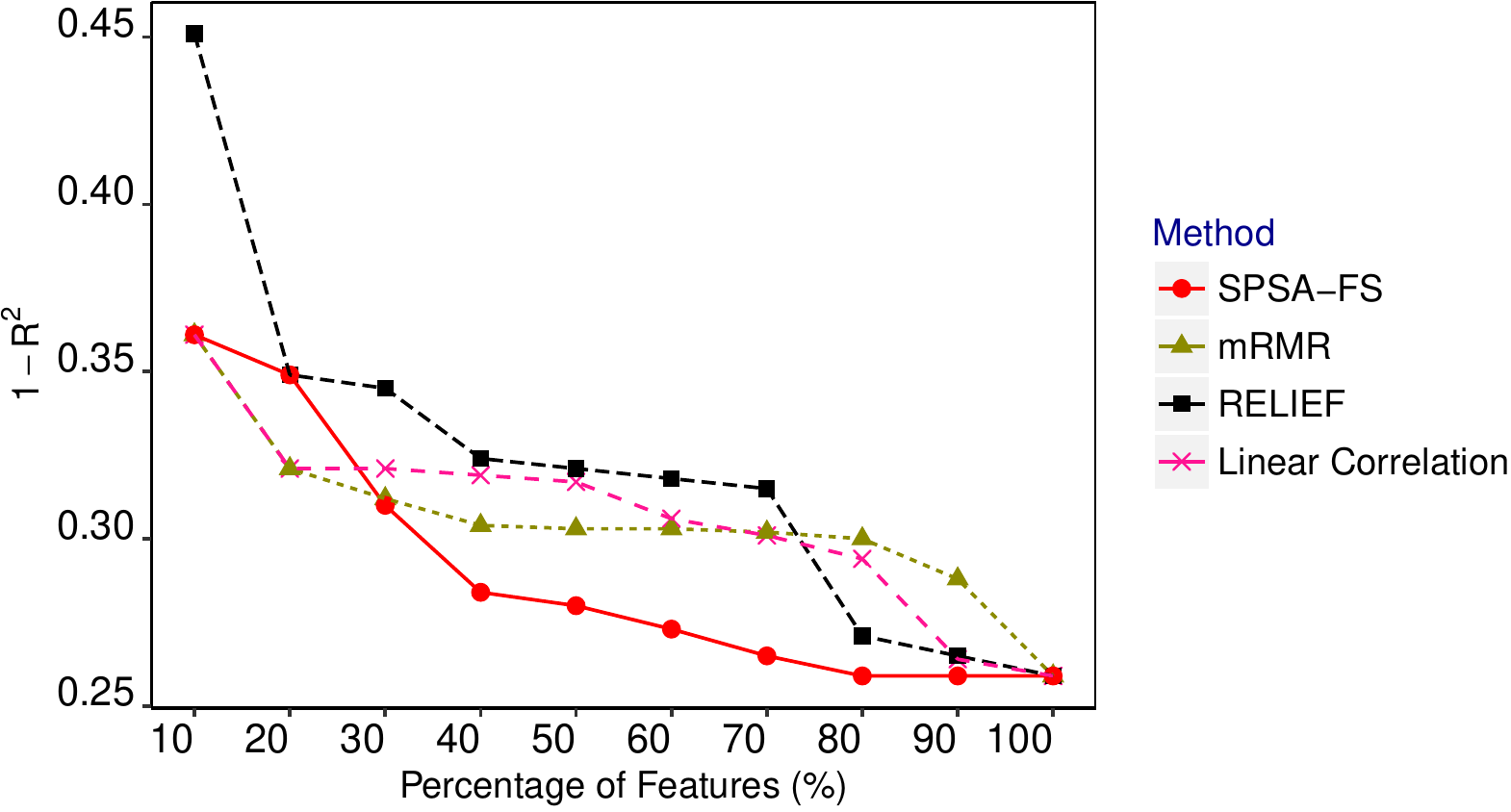}
\caption{\label{fig4}Regression on Boston Housing}
\end{figure}

\begin{figure}
\centering
\includegraphics{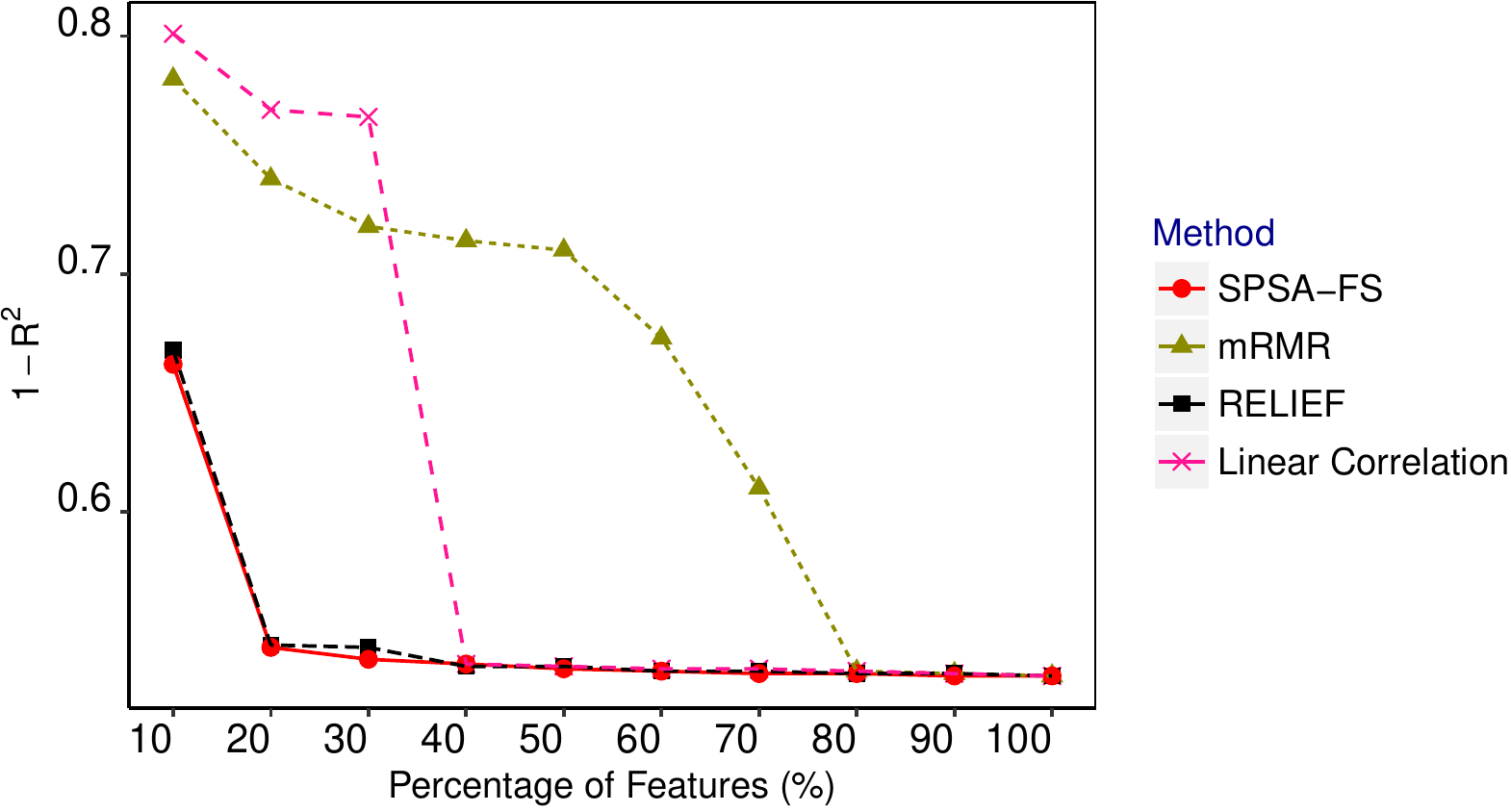}
\caption{\label{fig5}Regression on Pole Telecom}
\end{figure}

\begin{figure}
\centering
\includegraphics{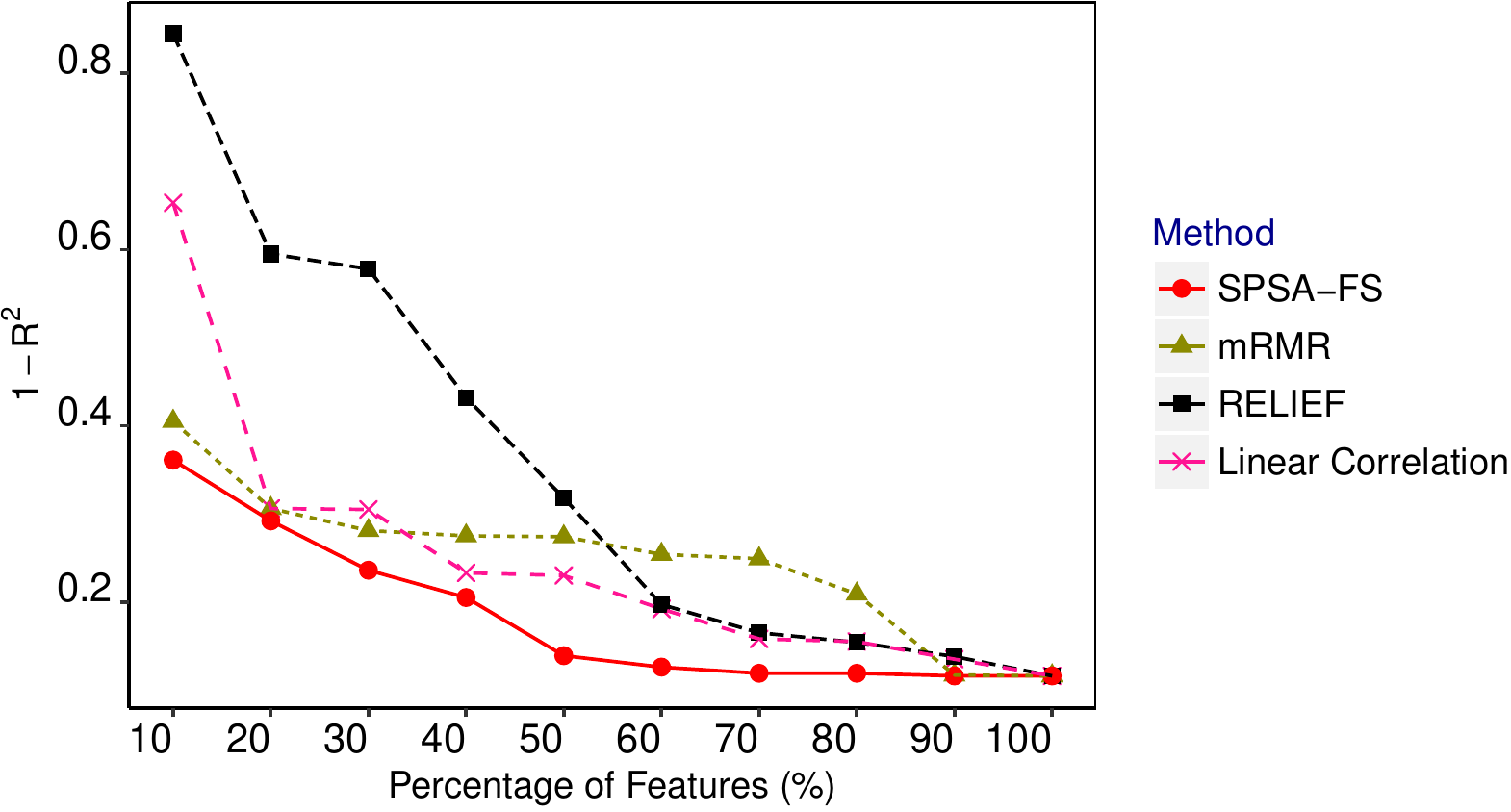}
\caption{\label{fig6}Regression on Pyrim}
\end{figure}

\begin{figure}
\centering
\includegraphics{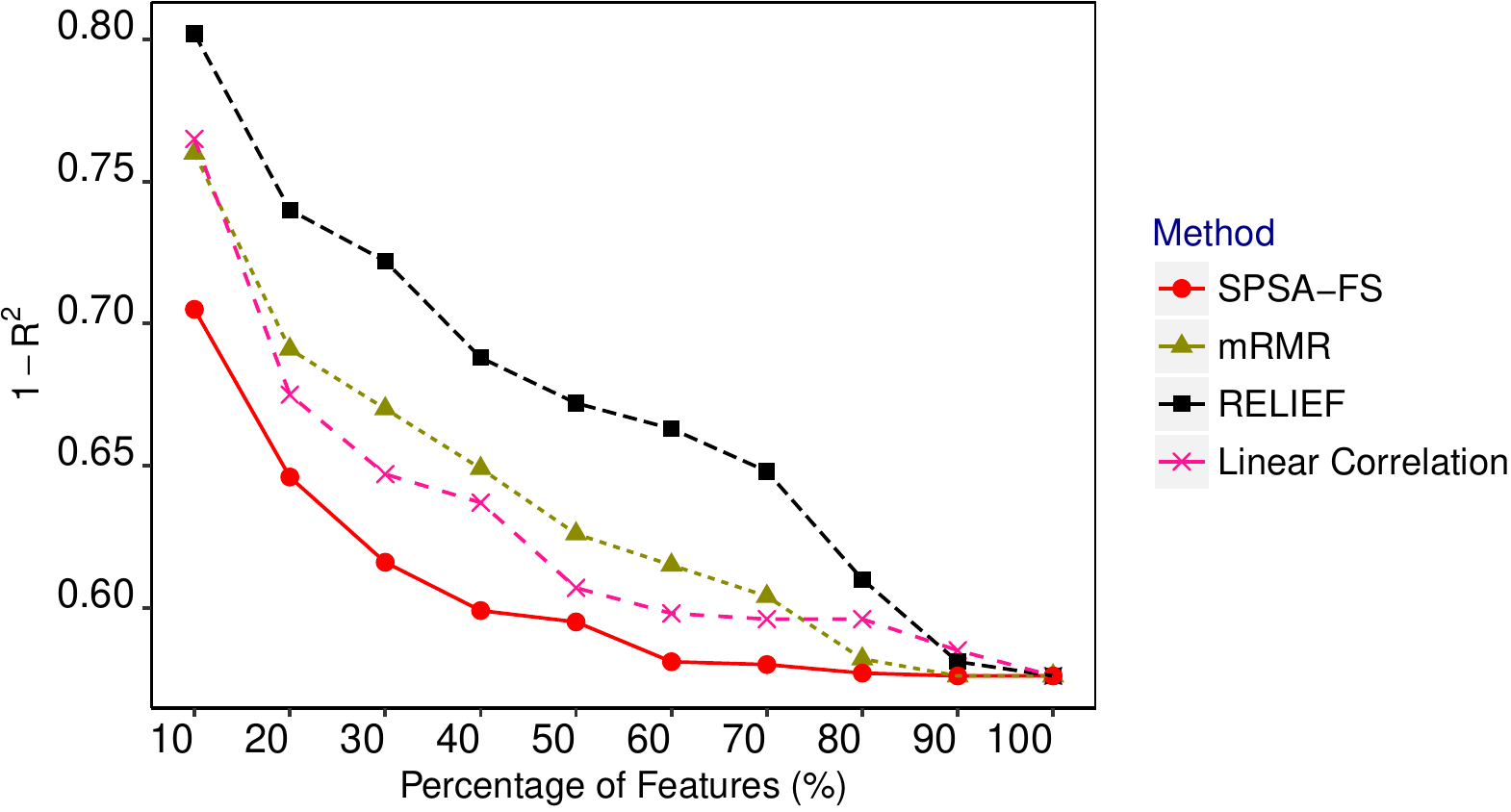}
\caption{\label{fig7}Regression on Triazines}
\end{figure}

\begin{figure}
\centering
\includegraphics{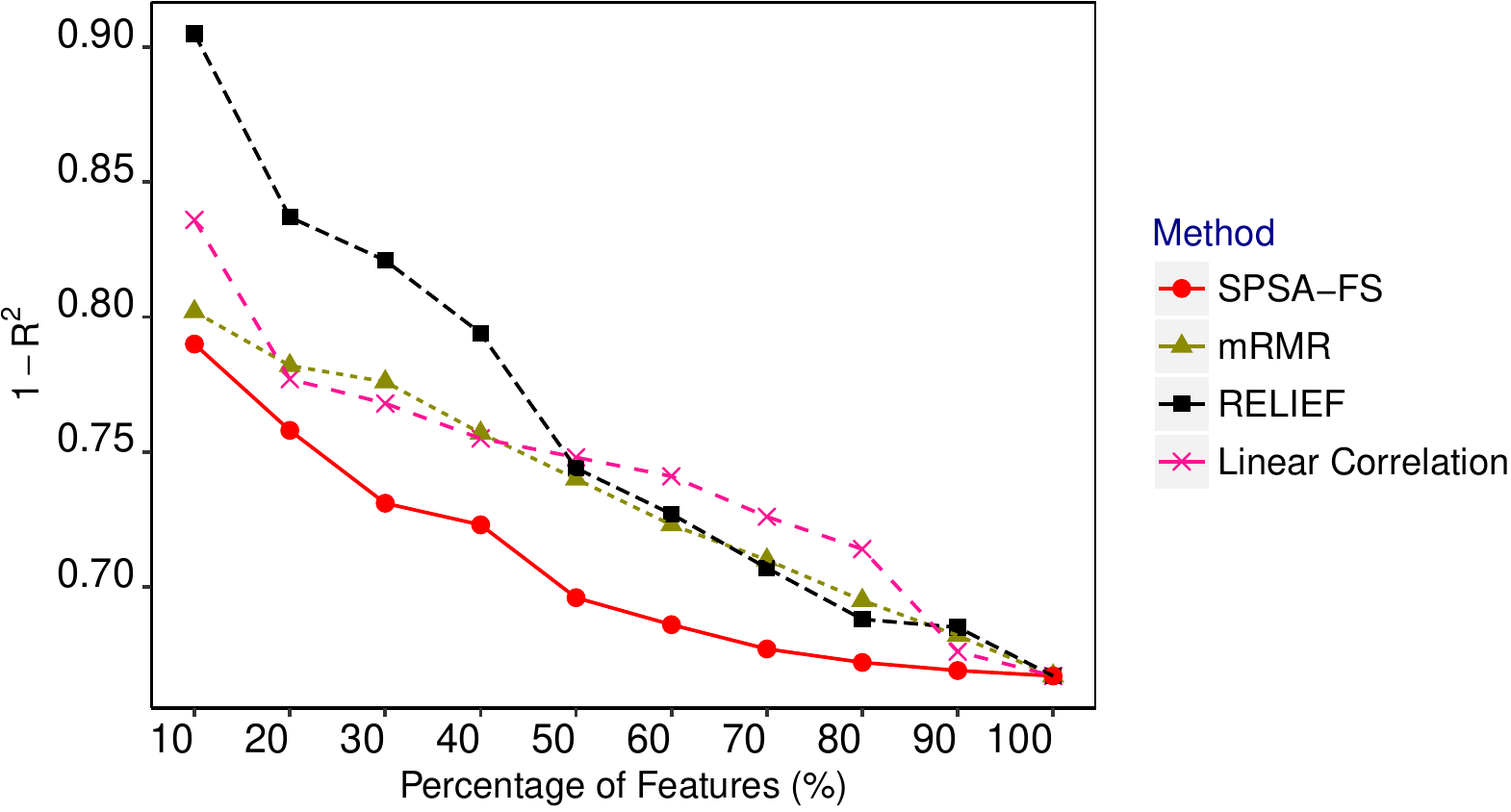}
\caption{\label{fig8}Regression on Wisconsin Breast Cancer}
\end{figure}

\chapter{Summary and Conclusions}\label{section5}

In this study, we propose the SPSA-FS algorithm which mitigates the slow
convergence issue of the BSPSA algorithm in feature selection. By
applying BB method to smooth step size and average gradient estimates,
SPSA-FS results in significantly lower computational costs but a very
minimal loss of accuracy. To vindicate our proposition, we ran
experiments which compared SPSA-FS to other wrapper methods on various
open datasets. For classification tasks, we evaluated accuracy
performance of the wrappers using the misclassification error rate. The
results were mixed since SPSA-FS's performance relied on the choice of
the classifier. However, SPSA-FS managed to strike a balance between
runtime required and accuracy. In the situation where SPSA-FS
outperformed, it yielded significantly higher accuracy rate. Meanwhile,
in the scenario where it underperformed, the performance differences
were marginal. For regression tasks, by using one minus R-squared as the
error measure, and SPSA-FS outperformed other wrappers with fewer
explanatory features. In conclusion, theoretically and empirically,
SPSA-FS not only leverages on the design of BSPSA which yields optimal
feature selection results but also gains substantial speed in locating
the solutions.

\chapter*{References}\label{references}
\addcontentsline{toc}{chapter}{References}

\hypertarget{refs}{}
\hypertarget{ref-vural}{}
Aksakalli, Vural, and Milad Malekipirbazari. 2016. ``Feature Selection
via Binary Simultaneous Perturbation Stochastic Approximation.''
\emph{Pattern Recognition Letters} 75 (Supplement C): 41--47.
doi:\href{https://doi.org/https://doi.org/10.1016/j.patrec.2016.03.002}{https://doi.org/10.1016/j.patrec.2016.03.002}.

\hypertarget{ref-ani}{}
Al-Ani, Ahmed. 2005. ``Feature Subset Selection Using Ant Colony
Optimization.'' \emph{International Journal of Computational
Intelligence} 2 (January): 53--58.

\hypertarget{ref-apolloni}{}
Apolloni, Javier, Guillermo Leguizamón, and Enrique Alba. 2016. ``Two
Hybrid Wrapper-Filter Feature Selection Algorithms Applied to
High-Dimensional Microarray Experiments.'' \emph{Applied Soft Computing}
38: 922--32.

\hypertarget{ref-bb}{}
Barzilai, J., and J. Borwein. 1988. ``Two-Point Step Size Gradient
Methods.'' \emph{IMA Journal of Numerical Analysis} 8: 141--48.

\hypertarget{ref-bennasar}{}
Bennasar, Mohamed, Yulia Hicks, and Rossitza Setchi. 2015. ``Feature
Selection Using Joint Mutual Information Maximisation.'' \emph{Expert
Systems with Applications} 42: 8520--32.

\hypertarget{ref-cadenas}{}
Cadenas, Jose M., M. Carmen Garrido, and Raquel Martinez. 2013.
``Feature Subset Selection Filter--Wrapper Based on Low Quality Data.''
\emph{Expert Systems with Applications} 40: 6241--52.

\hypertarget{ref-cauchy}{}
Cauchy, M. Augustine. 1847. ``Méthode Générale Pour La Résolution Des
Systèmes d'équations Simultanées.'' \emph{Comptes Rendus Hebd. Seances
Acad. Sci.} 25: 536--38.

\hypertarget{ref-chen}{}
Chen, Yu-Peng, Ying Li, Gang Wang, Yue-Feng Zheng, Qian Xu, Jia-Hao Fan,
and Xue-Ting Cui. 2017. ``A Novel Bacterial Foraging Optimization
Algorithm for Feature Selection.'' \emph{Expert Systems with
Applications} 83: 1--17.

\hypertarget{ref-dai2}{}
Dai, Y., and L. Liao. 2002. ``R-Linear Convergence of the Barzilai and
Borwein Gradient Method.'' \emph{IMA Journal of Numerical Analysis} 22:
1--10.

\hypertarget{ref-dai}{}
Dai, Y., W. Hager, K. Schittkowski, and H. Zhang. 2006. ``The Cyclic
Barzilai-Borwein Method for Unconstrained Optimization.'' \emph{Journal
of Numerical Analysis} 26: 604--27.

\hypertarget{ref-debuse}{}
Debuse, J. C. W., and V. J. Rayward-Smith. 1997. ``Feature Subset
Selection Within a Simulated Annealing Data Mining Algorithm.''
\emph{Journal of Intelligent Information Systems} 9 (January): 57--81.

\hypertarget{ref-mrmr}{}
Ding, C., and H. Peng. 2003. ``Minimum Redundancy Feature Selection from
Microarray Gene Expression Data.'' In \emph{Computational Systems
Bioinformatics. Csb2003. Proceedings of the 2003 Ieee Bioinformatics
Conference. Csb2003}, 523--28.
doi:\href{https://doi.org/10.1109/CSB.2003.1227396}{10.1109/CSB.2003.1227396}.

\hypertarget{ref-ghaemi}{}
Ghaemi, Manizheh, and Mohammed-Reza Feizi-Derakhshi. 2016. ``Feature
Selection Using Forest Optimization Algorithm.'' \emph{Pattern
Recognition} 60: 121--29.

\hypertarget{ref-guvenir1997supervised}{}
Guvenir, H A, B Acar, G Demiroz, and A Cekin. 1997. ``A Supervised
Machine Learning Algorithm for Arrhythmia Analysis.'' In \emph{Computers
in Cardiology}, 433--36.

\hypertarget{ref-guyon2}{}
Guyon, I., and Andre Elisseeff. 2003. ``An Introduction to Variable and
Feature Selection.'' \emph{Journal of Machine Learning Research} 3:
1157--82.

\hypertarget{ref-guyon}{}
Guyon, I., J. Weston, S. Barnhill, and V. Vapnik. 2002. ``Gene Selection
for Cancer Classifica- Tion Using Support Vector Machines.''
\emph{Machine Learning} 46: 389--422.

\hypertarget{ref-hsu}{}
Hsu, Hui-Huang, Cheng-Wei Hsieh, and Lu Ming-Da. 2011. ``Hybrid Feature
Selection by Combining Filters and Wrappers.'' \emph{Expert Systems with
Applications} 38: 8144--50.

\hypertarget{ref-relief}{}
Kira, Kenji, and Larry A. Rendell. 1992. ``The Feature Selection
Problem: Traditional Methods and a New Algorthm.'' In \emph{AAAI-92
Proceedings}.

\hypertarget{ref-kohavi}{}
Kohavi, R., and G. H. John. 1997. ``Wrappers for Feature Subset
Selection.'' \emph{Artificial Intelligence} 97 (1-2): 273--324.

\hypertarget{ref-rf}{}
Leo, B. 2001. ``Random Forests.'' \emph{Machine Learning} 45: 5--32.

\hypertarget{ref-ASU}{}
Li, Jundong, Kewei Cheng, Suhang Wang, Fred Morstatter, Trevino Robert,
Jiliang Tang, and Huan Liu. 2016. ``Feature Selection: A Data
Perspective.'' \emph{arXiv:1601.07996}.

\hypertarget{ref-UCI}{}
Lichman, M. 2013. ``UCI Machine Learning Repository.'' University of
California, Irvine, School of Information; Computer Sciences.
\url{http://archive.ics.uci.edu/ml}.

\hypertarget{ref-lu}{}
Lu, Huijuan, Junying Chen, Ke Yan, Qun Jin, Yu Xue, and Zhigang Gao.
2017. ``A Hybrid Feature Selection Algorithm for Gene Expression Data
Classification.'' \emph{Neurocomputing} 256: 56--62.

\hypertarget{ref-mafarja}{}
Mafarja, Majdi M., and Seyedali Mirjalili. 2017. ``Hybrid Whale
Optimization Algorithm with Simulated Annealing for Feature Selection.''
\emph{Neurocomputing} 260: 302--12.

\hypertarget{ref-molina}{}
Molina, B, and M Raydan. 1996. ``Preconditioned Barzilai-Borwein Method
for the Numerical Solution of Partial Differential Equations.''
\emph{Numerical Algorithms} 13: 45--60.

\hypertarget{ref-no}{}
Nocedal, Jorge, and Stephen J. Wright. 2006. \emph{Numerical
Optimization}. 2nd ed. Newyork: Springer.

\hypertarget{ref-oluleye}{}
Oluleye, B., L. Armstrong, and D. Diepeveen. 2014. ``A Genetic
Algorithm-Based Feature Selectionture Selection.'' \emph{International
Journal of Electronics Communication and Computer Engineering} 5
(April): 2278--4209.

\hypertarget{ref-olafsson}{}
Ólafsson, Sigurdur, and Jaekyung Yang. 2005. ``Intelligent Partitioning
for Feature Selection.'' \emph{INFORMS Journal on Computing} 17 (3):
339--55.

\hypertarget{ref-aydin}{}
Pashaei, Elnaz, and Nizamettin Aydin. 2017. ``Binary Black Hole
Algorithm for Feature Selection and Classification on Biological Data.''
\emph{Applied Soft Computing} 56: 94--106.

\hypertarget{ref-peng}{}
Peng, H., F. Long, and C. Ding. 2005. ``Feature Selection Based on
Mutual Information: Criteria of Max-Dependency, Max-Relevance, and
Min-Redundancy.'' \emph{IEEE Transactions on Pattern Analysis and
Machine Intelligence}, no. 1226--1238.

\hypertarget{ref-pudil}{}
Pudil, P., J. Novovicová, and J. Kittler. 1994. ``Floating Search
Methods in Feature Selection.'' \emph{Pattern Recognition Letters} 15
(October): 1119--25.

\hypertarget{ref-R}{}
R Core Team. 2017. \emph{R: A Language and Environment for Statistical
Computing}. Vienna, Austria: R Foundation for Statistical Computing.
\url{https://www.R-project.org/}.

\hypertarget{ref-raydan}{}
Raydan, M. 1993. ``On the Barzilai and Borwein Choice of Steplength for
the Gradient and Method.'' \emph{IMA Journal of Numerical Analysis} 13:
321--26.

\hypertarget{ref-raydan2}{}
Raydan, M, and B. Svaiter. 2002. ``Relaxed Steepest Descent and
Cauchy-Barzilai-Borwein Method.'' \emph{Computational Optimization and
Applications} 21: 155--67.

\hypertarget{ref-raymer}{}
Raymer, Michael L., William F. Punch, Erik D. Goodman, Leslie Kuhn, Anil
K. Jain, and et al. 2000. ``Dimensionality Reduction Using Genetic
Algorithms.'' \emph{Evolutionary Computation, IEEE Transactions on} 4
(February): 164--71.

\hypertarget{ref-sayed}{}
Sayed, Safinaz AbdEl-Fattah, Emad Nabil, and Amr Badr. 2016. ``A Binary
Clonal Flower Pollination Algorithm for Feature Selection.''
\emph{Pattern Recognition Letters} 77: 21--27.

\hypertarget{ref-senawi}{}
Senawi, Azlyna, Hua-Liang Wei, and Stephan A. Billings. 2017. ``A New
Maximum Relevance-Minimum Multicollinearity (Mrmmc) Method for Feature
Selection and Ranking.'' \emph{Pattern Recognition} 67: 47--61.

\hypertarget{ref-sikonia}{}
Sikonia, M. R., and I. Kononenko. 2003. ``Theoretical and Empirical
Analysis of Relief and Relieff.'' \emph{Machine Learning} 53 (23-69).

\hypertarget{ref-spall}{}
Spall, James C. 1992. ``Multivariate Stochastic Approximation Using a
Simultaneous Perturbation Gradient Approximation.'' \emph{IEEE} 37 (3):
322--41.

\hypertarget{ref-spall2}{}
Spall, James C. 2003. \emph{'Introduction to Stochastic Search and
Optimization: Estimation, Simulation, and Control'}. John Wiley.

\hypertarget{ref-wang}{}
Spall, James C., and Wang Qi. 2011. ``Discrete Simultaneous Perturbation
Stochastic Approximation on Loss Function with Noisy Measurements.''
\emph{'In: Proceeding American Control Conference'} 37 (3): 4520--5.

\hypertarget{ref-tahir}{}
Tahir, M. A., A. Bouridane, and F. Kurugollu. 2007. ``Simultaneous
Feature Selection and Feature Weighting Using Hybrid Tabu
Search/K-Nearest Neighbor Classifier.'' \emph{Pattern Recognition
Letters} 28 (April): 438--46.

\hypertarget{ref-tan}{}
Tan, Conghui, Shiqian Ma, Yu-Hong Dai, and Yuqiu Qian. 2016.
``Barzilai-Borwein Step Size for Stochastic Gradient Descent.''
Barcelona.

\hypertarget{ref-lasso}{}
Tibshirani, Robert. 1996. ``Regression Shrinkage and Selection via the
Lasso.'' \emph{Journal of the Royal Statistical Society, Series B} 58:
267--88.

\hypertarget{ref-DCC}{}
Torgo, L. 2017. ``DCC Regression Datasets.'' Universidade Do Porto,
Portugal, Department of Computer Science.
\url{http://www.dcc.fc.up.pt/~ltorgo/Regression/DataSets.html}.

\hypertarget{ref-tsai}{}
Tsai, Chih-Fong, William Eberle, and Chi-Yuan Chu. 2013. ``Genetic
Algorithms in Feature and Instance Selection.'' \emph{Knowledge-Based
Systems} 39: 240--47.

\hypertarget{ref-wan}{}
Wan, Youchuan, Mingwei Wang, Zhiwei Ye, and Xudong Lai. 2006. ``A
Feature Selection Method Based on Modified Binary Coded Ant Colony
Optimization Algorithm.'' \emph{Applied Soft Computing} 49: 248--58.

\hypertarget{ref-wang2}{}
Wang, Lipo, Yaoli Wang, and Qing Chang. 2016. ``Feature Selection
Methods for Big Data Bioinformatics: A Survey from the Search
Perspective.'' \emph{Methods} 111 (December): 21--31.

\hypertarget{ref-wang3}{}
Wang, X., J. Yang, X. Teng, W Xia, and R. Jensen. 2007. ``Feature
Selection Based on Rough Sets and Particle Swarm Optimization.''
\emph{Pattern Recognition Letters} 28 (April): 459--71.

\hypertarget{ref-hadley}{}
Wickham, Hadley. 2016. \emph{Ggplot2: Elegant Graphics for Data
Analysis}. Springer-Verlag New York. \url{http://ggplot2.org}.

\hypertarget{ref-yihui}{}
Xie, Yihui. 2015. \emph{Dynamic Documents with R and Knitr}. 2nd ed.
Boca Raton, Florida: Chapman; Hall/CRC. \url{https://yihui.name/knitr/}.

\hypertarget{ref-iso}{}
Zheng, Zhonglong, Xie Chenmao, and Jiong Jia. 2010. ``ISO-Container
Projection for Feature Extraction.'' IEEE.

\end{document}